\documentclass[journal]{IEEEtran}
\usepackage{amsmath,amssymb,amsfonts}

\usepackage{epsfig}
\usepackage{subfig}

\usepackage{threeparttable}
\usepackage{booktabs}
\usepackage{url}
\usepackage{multirow}

\usepackage{fixmath}
\usepackage{graphicx}
\usepackage{float}

\usepackage{algorithm}
\usepackage{algpseudocode}

\usepackage{enumitem}

\graphicspath{{./figures/}}

\def\({\left(}
\def\){\right)}
\def\[{\left[}
\def\]{\right]}
\def\nn{\nonumber}

\def\bx{\boldsymbol{x}}
\def\bX{\boldsymbol{X}}
\def\hbx{\widehat{\bx}}
\def\btheta{\boldsymbol{\theta}}
\def\hbtheta{\widehat{\boldsymbol{\theta}}}

\def\bu{\boldsymbol{u}}
\def\bbu{\bar{\boldsymbol{u}}}
\def\bU{\boldsymbol{U}}
\def\bbU{\bar{\boldsymbol{U}}}
\def\bUst{\bU^\star}

\def\bV{\boldsymbol{V}}
\def\bv{\boldsymbol{v}}
\def\bbv{\bar{\bv}}
\def\bvst{\bv^\star}

\def\bQ{\boldsymbol{Q}}
\def\bp{\boldsymbol{p}}
\def\bP{\boldsymbol{P}}

\def\bc{\boldsymbol{c}}

\def\bW{\boldsymbol{W}}
\def\bM{\boldsymbol{M}}
\def\br{\boldsymbol{r}}

\def\E{\mathbb{E}}
\def\L{\mathcal{L}}

\def\bI{\boldsymbol{I}}
\def\b0{\boldsymbol{0}}

\def\sI{\mathcal{I}}
\def\bPsi{\boldsymbol{\Psi}}

\def\eps{\epsilon}

\def\lam{\lambda}
\def\lamst{\lambda^\star}
\def\blam{\boldsymbol{\lambda}}

\def\bSigma{\boldsymbol{\Sigma}}

\def\tr{\text{tr}}

\def\bn{\boldsymbol{\eta}}
\newcommand{\bna}[2]{\boldsymbol{\eta}_{#1,#2}}

\newcommand{\paren}[1]{\left(#1\right)}

\begin{document}
\title{Online Forecasting Matrix Factorization}
\author{San Gultekin and John Paisley}
\maketitle

\begin{abstract}
In this paper the problem of forecasting high dimensional time series is considered. Such time series can be modeled as matrices where each column denotes a measurement. In addition, when missing values are present, low rank matrix factorization approaches are suitable for predicting future values. This paper formally defines and analyzes the forecasting problem in the online setting, i.e. where the data arrives as a stream and only a single pass is allowed. We present and analyze novel matrix factorization techniques which can learn low-dimensional embeddings effectively in an online manner. Based on these embeddings a recursive minimum mean square error estimator is derived, which learns an autoregressive model on them. Experiments with two real datasets with tens of millions of measurements show the benefits of the proposed approach.
\end{abstract}


\section{Introduction}
\label{sec:intro}

A brief look into the history of signal processing reveals an ever-increasing need for acquiring, storing, and analyzing data which constantly changes in form and presents new challenges. In the big data era, the challenge has been growing in different dimensions. On one hand, many techniques become obsolete as the scale of the data is witnessing unprecedented growth. On the other hand, the complexity of data is increasing. Indeed, modern datasets are composed of many different features that can include, for example, text, audio, image as well as various numerical and categorical features. Another very important issue is the case of missing data, which can have multiple causes; e.g. corruption during acquisition, or the data-specific sparsity. For instance, an online retailer will have many customers and products, but each customer will be interacting with a very small subset of all products, giving rise to a very sparse interaction history.

Collaborative filtering is one particular field in which one is typically required to deal with missing values. Here, the aim is to recommend a user a set of items, based on other users with ``similar" interests. While there are a number of ways to address this problem, matrix factorization has been one of the most successful approaches to solve this problem, being the winner of one million dollar Netflix challenge \cite{Koren_2009}. The approach taken here is to treat the entire data as a matrix with many missing entries, and factor it as a product of two low rank matrices. These factors can then be used to predict missing entries. Various matrix factorization algorithms has been successfully applied, in many different settings in addition to the collaborative filtering \cite{Salakh_2008a, Salakh_2008b}. Some examples include natural language processing \cite{Falahatgar_2016, Pennington_2014}, image processing \cite{Bach_2010}, and power systems analysis \cite{Mei_2017}. In addition to these, the special case of nonlinear matrix factorization has also received significant attention \cite{Seung_2001}.

While matrix factorization has been a popular choice for many different problems, interestingly, its applications to time series analysis has been less developed. Modern time series are typically high dimensional, with many missing values; consequently, the entire time series can be treated as a sparse matrix, and low-rank matrix factorization can be quite useful. The recent work of \cite{Dhillon_2016} proposed a temporal regularized matrix factorization, based on this observation. A key property of their solution is that, the columns of one of the factor matrices is regularized by an AR process. The coefficients of this process are learned from the data, and can be used to forecast future values. This shows that matrix factorization is also a powerful tool for time series forecasting. With that said, the emphasis in \cite{Dhillon_2016} was on the batch learning case. For many practical applications, it might be impractical to load and process the entire batch, or the data itself may be arriving as a stream. This brings one to the challenging setting of online learning, where the data is processed as a stream, and no storage or multiple passes are allowed. Online learning for time series prediction is an active area of research \cite{Anava_2013, Liu_2016, Koolen_2015}. In particular, the recent work \cite{Anava_2015} considers online predictions with missing values. However, to the best of our knowledge, online forecasting of time series with matrix factorization has not been considered before. In this paper we formally introduce the problem and propose novel techniques, which forecast future values of high dimensional time series, based on matrix factorization. A key observation in previous works \cite{Anava_2013, Anava_2015} is that, the textbook methods to time series analysis which typically assume stationarity of series and/or Gaussianity of noise terms are oftern unrealistic: This is also our preferred approach here and throughout the paper we make minimal assumptions about the data generating process.

The problem we consider in this paper falls into the general framework of dynamic matrix factorization, in which one finds time-varying factors for a time-varying observation matrix. A particularly appealing approach to dynamic matrix factorization utilizes state-space models; here, the columns of latent factor matrices follow a generative state-space model, and their values are inferred from the observations. This is equivalent to a non-convex version of the Kalman filter \cite{Kalman_1960}. Several recent works consider dynamic, state-space models in batch setting \cite{Sun_2014, Mohammadiha_2015}. On the other hand, \cite{Gultekin_2014} proposed an online, dynamic state-space model, named Collaborative Kalman Filter (CKF), which can estimate the states in an online manner. It is also worthwhile to note that many well established algorithms such as probabilistic matrix factorization \cite{Salakh_2008a} can also be extended to the online and dynamic setting. Another recent work \cite{Xu_2016} is concerned with providing theoretical guarantees in the dynamic setting. With that said, our problem setting is different from the previous literature in two senses: First, in the online setting we no longer observe a dynamic matrix at each time, but instead we observe a single column, i.e. cross section of a high dimensional time-series. This is a degenerate setting, as a naive matrix factorization approach will always give rank-1 approximations. Secondly, while the previous work focuses on predicting missing entries of the current observation matrix, the forecasting problem is concerned with predicting future values, for which an additional extrapolation step is necessary.

Finally we note that, there is a significant body of work which consider the missing value and forecasting problems with other approaches: \cite{Fanghan_2013} proposes a convex optimization framework for transition matrix estimation in vector-valued time series. While \cite{Dunsmuir_1981} employs a time-series model to represent missing observations, \cite{Shumway_1982} handles them with an EM algorithm. The work \cite{Choong_2009} utilizes an AR process to impute missing values, and \cite{Sinopoli_2004} considers Kalman filtering problem with intermittent observations. The main benefit of of using matrix factorization is that, the low rank representation is a natural choice for high dimensional and sparse time series, and as shown in this paper, non-trivial low rank factorizations can be learned efficiently in the online setting.

We organize this paper as follows: Section \ref{sec:background} establishes the background for AR processes and matrix factorization approach to time series analysis. Section \ref{sec:mf} is concerned with introducing matrix factorization methods, which find low-rank factorizations suitable for forecasting. Building on such factorization, Section \ref{sec:pred} shows how the coefficients of the AR process can be estimated in an optimum manner. Section \ref{sec:exp} contains extensive experimentation with two real datasets with tens on millions of measurements; our experiments show that the proposed techniques are effective in practical situations. We conclude in Section 6.

\section{Background}
\label{sec:background}

This section provides background on time series and matrix factorization, and introduces a generative model which provides a framework for subsequent development. In this paper, we are interested in forecasting the future values of a high dimensional time series $\{\bx_t\}_{t=1}^T$, where each $\bx_t$ is an $M \times 1$ vector. In particular, the analysis will take place in the online setting where, at each time step $t$, the value of $\bx_t$ must be predicted - denoted by $\hbx_t$ - before it is observed, and upon observation the predictor suffers a loss and updates its model. For this paper, the time indices are discrete and homogeneous; the current time and total number of samples are denoted by $t$ and $T$, and also $[T] = \{1,2,\ldots,T\}$. In the well-known Box-Jenkins approach \cite{Brockwell_2016}, given the samples, one constructs a signal model by finding (i) a trend, (ii) a seasonal component, and (iii) a noise component, where the latter is typically modeled by an autoregressive moving average (ARMA) model, which is a combination of the AR and MA models. The learned model can then be assessed with appropriate statistical tests. One drawback of this approach is that, finding a trend and seasonal component requires storage and processing of the entire data, which might be unsuitable due to storage or computation time requirements. In addition, this methodology is also unsuitable for streaming data. 

For these reasons, we start from a generic vector AR process (VAR(P)) model of form
\begin{align} \label{eq:ar_model}
	\bx_t = \theta_1 \bx_{t-1} + \ldots + \theta_P \bx_{t-P} + \bna{\bx}{t} ~,	
\end{align}
where $\bna{\bx}{t}$ is zero mean white noise and $P$ denotes the model order. The choice of P has a major impact on the accuracy of the model, as it captures the maximal lag for correlation. Let the parameters of this model be denoted by $\btheta = [\theta_1,\ldots,\theta_P]^\top$ for shorthand. It is clear that, with scalar coefficients the VAR(P) corresponds to M copies of an AR(P) model. In addition, when the polynomial $\psi^P - \theta_1 \psi^{P-1} - \ldots -\theta_P$ has roots inside the unit circle, the model is stationary \cite{Hamilton_1994}. For a given \emph{finite} number of measurements, the parameters of the AR(P) model can be estimated by minimizing the mean square error
\begin{align}\label{eq:var}
	\hbtheta = \arg \min_{\btheta'} \E_{\theta} ~ \| \btheta - \btheta' \|_2^2 ~,
\end{align}
where expectation is taken with respect to the prior $p(\btheta)$. A particular advantage of this formulation is, the optimum linear minimum mean squared error estimator (LMMSE) does not make any distribution assumptions on $p(\btheta)$ or $p(\bn)$, and can be calculated in closed form, given the first and second order statistics. More importantly, unlike the least square or best linear unbiased (BLU) estimator, LMMSE estimator is guaranteed to exist. We will give a derivation of this estimator in Section 4, where we discuss optimum sequence prediction.

We now turn to matrix factorization, first note that the time series at hand can be represented by an $M \times T$ matrix $\bX$. Let $d$ denote the rank of this matrix; as a direct consequence of the full rank partitioning property \cite{Gentle_2007}, it is possible to find a $d \times M$ matrix $\bU$ and a $d \times T$  $\bV$ such that $\bX = \bU^\top \bV$. Note that such a factorization is non-unique and there are multiple ways to find one. For example singular value decomposition (SVD) is widely used for this purpose. Furthermore, when $\bX$ represents a time series, such a factorization can be interpreted as follows: Since the matrix $\bV$ is $d \times T$, it corresponds to a \emph{compression} of the original $M \times T$ matrix $\bX$; therefore the matrix $\bV$ is itself a time series; whereas the matrix $\bU$ provides the combination coefficients to reconstruct $\bX$ from $\bV$. Based on this key observation, \cite{Dhillon_2016} proposed a temporal regularized matrix factorization, where the regularizer on the columns of $\bV$ is in the form of an AR process; it is showed that such regularization has notable impact on performance.

\begin{figure*}[t]
	\includegraphics[width=\textwidth]{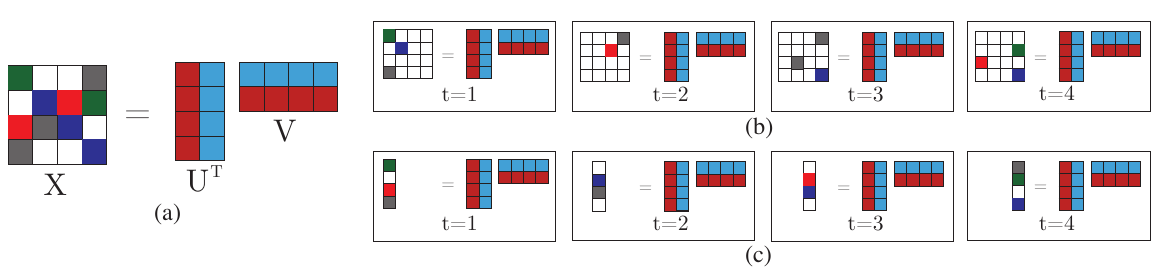}
	\caption{Comparison of online matrix factorization schemes. In panel (a), a matrix $\boldsymbol{X}$ is factorized in batch setting, whereas in panel (b), at each time, a subset of the matrix is observed. For illustrative purposes, note that the observed rank is always greater than one, so the problem is feasible. Panel (c) shows online matrix factorization, where at each time a vector is observed and factorization rank never exceeds one. This is the setting of the paper.}
		\label{fig:mf_schemes}
\end{figure*}

Motivated by this, our goal in this paper is to learn the factorizations $\bU$ and $\bV$, along with the AR model of Eq. \eqref{eq:var} within the \emph{online} setting, where, at each time instant, we observe a single column of the data matrix, namely $\bx_t$. While this has similarity to the previous work on online/dynamic matrix factorization algorithms \cite{Gultekin_2014,Sun_2014}, one fundamental issue sets it apart: In the aforementioned papers, at each time an $M \times N$ matrix is observed, with $N \gg 1$; whereas in our case the observation is simply $M \times 1$. This is illustrated in Figure \ref{fig:mf_schemes}, in panel (a) we show the batch factorization of an $M \times T$ matrix $\boldsymbol{X}$. Panel (b) is the case where, at each time, a subset of the matrix entries are observed. However, when $\bX$ is a time series matrix, at each time we observe a single column, as shown in panel (c).

The problem with observing a vector is, the latent rank is at most $1$ in this case; whereas the batch problem in Figure \ref{fig:mf_schemes}(a) will have a solution of rank $d$. Since the end goal here is to factor the entire data, it is desirable to start from a rank-$d$ representation and gradually update it. However, finding such factors naively gives poor performance (Figure 4), therefore our task is to devise an effective way achieving this. This can be done using specific penalties on the matrix $\bU$, which yields feasible optimization problems, as shown in the next section. One way to motivate our approach is to consider a probabilistic generative model for the data, as frequently used in Bayesian methods \cite{Bishop_2006}. Here we employ a state-space representation
\begin{align}\label{eq:gen}
	\bU_t &= \bU_{t-1} + \bna{\bU}{t} \nn \\
	\bv_t &= \theta_1 \bv_{t-1} + \ldots + \theta_P \bv_{t-P} + \bna{\bv}{t} \nn \\
	\bx_t &= \bU_t^\top \bv_t + \bna{\bx}{t} ~,
\end{align}
where $[\bna{\bU}{T}]$, $[\bna{\bv}{T}]$, and $[\bna{\bx}{T}]$ are white noise sequences, independent of each other. This model reveals, an EM-like algorithm is necessary: In the E-step $\bU_t$ and $\bv_t$ are estimated; in the M-step, treating the E-step estimates as observations, $\theta_1,\ldots,\theta_P$ are updated. These two steps are covered in Sections \ref{sec:mf} and \ref{sec:pred} respectively.

\section{ONLINE MATRIX FACTORIZATION}
\label{sec:mf}

The algorithms presented in the paper fit an AR model to the sequence $[\bv_T] = \{\bv_1,\ldots,\bv_T\}$, which itself is generated in an online fashion. To get a good fit, it is necessary to generate the vectors $\bv_t$ is a proper manner. To illustrate this, for a given measurement vector if we find a factorization $\bx_t = \bU_t^\top \bv_t$, for any orthogonal matrix $\bQ$ and positive scaling constant $a$ we get $\bx_t = (a\bQ\bU_t)^\top (a^{-1}\bQ\bv_t)$. This scaling and rotation could have a significant effect on forecasting accuracy (Figure 4). On the other hand, the matrix $\bU_t$ is a slowly time-varying quantity which means we can constrain its variation. Accurate selection of the penalty on $\bU_t$ has a dramatic effect on the generated $[\bv_T]$, which then dictates the forecasting accuracy.

Another mild assumption we use is that, the absolute value of observations are upper bounded by a finite number. Further, without loss of generality we impose $\sup_{\bx \in [\bx_T]} \|\bx\|_\infty = 1$, i.e. the upper bound is simply unity. This is not a restrictive assumption, as an upper limit on the measurements are typically known and normalization can be applied accordingly. For the datasets used in this paper, for the power data, this is dictated by the physical constraints of the network, i.e. capacity of the elements; for the traffic data, the measurements are in percentages, which automatically satisfy the bound.

\subsection{Fixed Penalty Constraint}
\label{sec:fp}

The first algorithm we present is based on a simple fixed penalty function on the norms of the factors. The batch version for this algorithm was previously considered in \cite{Salakh_2008a}. For that case the cost function is 
\begin{align}\label{eq:batch_pmf}
	f(\bU,\bv) = \sum_{m,n} (x_{m,n} - \bu_m^\top \bv_n)^2  + \rho_u \|\bU\|_F^2 + \rho_v \|\bV\|_F^2    
\end{align}
This is equivalent to adding Gaussian priors to each column of $\bU$ and $\bv$ and in \cite{Salakh_2008a} is referred to as probabilistic matrix factorization (PMF). This non-convex, unconstrained problem is solved by an alternating coordinate descent scheme
\begin{align}
	\bu_m^{(i)} &\gets \paren{ \rho_u \bI + \sum_n \bv_n^{(i)} \bv_n^{(i)\top} }^{-1} \paren{ \sum_n x_{m,n} \bv_n^{(i)} } \nn \\
	\bv_n^{(i+1)} &\gets \paren{ \rho_v \bI + \sum_m \bu_m^{(i)} \bu_m^{(i)\top} }^{-1} \paren{ \sum_m x_{m,n} \bu_m^{(i)} }
\end{align}
These updates are easily found by matrix differentiation. Now turning to the online case, at each time, a single column of $\bX$ is revealed. Imposing the generative model of Eq. \eqref{eq:gen}, at time $t$ we would like to minimize the following cost function
\begin{align}\label{eq:fp}
	f(\bU_t,\bv_t) = \|\bx_t - \bU_t^\top \bv_t\|_2^2  + \rho_u \|\bU_t - \bbu\|_F^2 + \rho_v \|\bv_t - \bbv\|_2^2    
\end{align}

As a subtlety, note that $\bU_t$ will be the submatrix of $\bU$ which contains the columns with a corresponding observation in $\bx_t$; so when there are missing observations, only a subset of columns get updated. At this point it is important to make a distinction between the update for $\bU_t$ and $\bv_t$. At any given time, the number of observations is $M_t$, and $\bv_t$ has $d$ parameters. Typically $M_t>d$ and the update for $\bv_t$ can be feasible even if $\rho_v = 0$; therefore $\rho_v$ is a small set-and-forget constant that we include for numerical stability \footnote{Indeed, $\rho_v = 10^{-4}$ for all experiments in this paper.}. On the other hand, $\bU_t$ contains $M_t d > M_t$ unknowns and the Gram matrix $\bU_t \bU_t^\top$ is not invertible. Therefore, $\rho_u>0$ is necessary to make the problem feasible. Since both $\rho_u$ and $\rho_v$ are fixed at the beginning, we refer to Eq. \eqref{eq:fp} as fixed penalty (FP) matrix factorization. This naming choice will be evident in the next section.

Comparing the generative model of Eq. \eqref{eq:gen} and the cost function of Eq. \eqref{eq:fp}, FP finds the maximum a posteriori (MAP) solution. Here, the priors follow state equations: $\bbU = \bU_{t-1}$  and $\bbv = \sum_{l=1}^P \theta_l \bv_{t-l}$. Moreover, the $\ell_2$-norm terms in Eq. \eqref{eq:fp} suggests that $\bna{\cdot}{t}$ has a density inversely proportional to the distance from the mean. Indeed, this is the only assumption we make about the noise p.d.f. While the most common choice satisfying this requirement would be the Gaussian density; note that its support is the entire $\mathbb{R}^n$ which conflicts with the bounded data assumption. Secondly, from the perspective of Eq. \eqref{eq:gen}, the regularization coefficients can be regarded as the inverse noise variance; higher values mean higher trust to the prior and stronger regularization. Finally, we note that whereas $\rho_v$ is usually a small constant, $\rho_u \gg \rho_v$. This suggests, at time $t$, FP finds a solution for which $\bU_t$ remains close to $\bU_{t-1}$, i.e. $\bU_t$ is slowly time-varying. This agrees with the interpretation that, in the batch case $\bU$ is a fixed set of coefficients and $\bV$ contains the compressed time series. Another caution here is that, setting $\rho_v$ high would over-constrain the problem as both $\bU_t$ and $\bv_t$ would be forced to stay close to $\bbU$ and $\bbv$ while trying to minimize the approximation error to $\bx_t$.

The update equations for FP are
\begin{align}\label{eq:fp_update}
	\bU_t^{(i)} &\gets (\rho_u \bI + \bv_t^{(i)} \bv_t^{(i)\top})^{-1} (\rho_u \bbU + \bv_t^{(i)} \bx_t^\top) \nn \\
	\bv_t^{(i)} &\gets (\rho_v \bI + \bU_t^{(i)} \bU_t^{(i)\top})^{-1} (\rho_v \bbv + \bU_t^{(i)} \bx_t) ~.
\end{align}

A key argument in Eq. \eqref{eq:fp} is that, the state equations of Eq. \eqref{eq:gen} addresses scaling and rotation issues through $\bbU$ and $\bbv$. A naive approach, which does not impose any temporal structure on the latent variables, constructs the alternative objective
\begin{align}\label{eq:naive_mf}
	f(\bU_t,
	\bv_t) = \|\bx_t - \bU_t^\top \bv_t \|_2^2 + \rho_u \|\bU_t\|_F^2 + \rho_v \|\bv_t\|_2^2 ~.   
\end{align} 
We also consider this alternative ``naive" model in the experiments, to show that, in the absence of temporal regularization in Eq. \eqref{eq:gen}, scaling and rotation cannot be prevented \footnote{One alternative way to address this problem would utilize post-processing. In particular, the optimum rotation between two sets of points can be found by solving the Procrustes problem \cite{Horn_2012} however this would incur additional computation.}, hindering the prediction quality. The FP matrix factorization is summarized in Algorithm \ref{alg:fp}.

\begin{algorithm}[t]
\caption{Fixed Penalty Matrix Factorization (FP)}\label{alg:fp}
\begin{algorithmic}[1]
\State \textbf{Require: } $\bx_t$, $\sI_t$, $\rho_u$, $\rho_v$, $\bbU$, $\bbv$, ${\tt max\_ite}$
\State \textbf{Return: $\bU_t$, $\bv_t$}
\vspace{.05in}
\State Re-assign $\bbU \gets \bbU(:,\sI_t)$
\For{$i=1,\dots,{\tt max\_ite}$}
\vspace{.05in}
\State $\bv^{(i)} \gets (\rho_v \bI + \bU^{(i-1)} \bU^{(i-1)\top})^{-1} (\rho_v \bbv + \bU^{(i-1)} \bx_t)$
\State $\bU^{(i)} \gets (\rho_u \bI + \bv^{(i)} \bv^{(i)\top})^{-1} (\rho_u \bbU + \bv^{(i)} \bx_t^\top)$
\EndFor
\State Update $\bU_t(:,\sI_t) \gets \bU$ and $\bU_t(:,\sI_t^c) \gets \bU_{t-1}(:,\sI_t^c)$.
\State Update $\bv_t \gets \bv$.
\vspace{.1in}
\State Note 1: We use Matlab-like slicing to denote columns. In particular $\bU_t(:,\sI_t)$ are those columns for which there is a corresponding observation at time $t$. $\bU_t(:,\sI_t^c)$ is simply the remaining columns.
\State Note 2: Based on this, at a given time t, only the observed entries get updated, and their value at $t-1$ is added as Frobenius norm constraint. 
\end{algorithmic} 
\end{algorithm}

\subsection{Fixed Tolerance Constraint}
\label{sec:ft}

The fixed penalty approach to matrix factorization suffers from several potential issues: While $\rho_v$ is a small set-and-forget constant, picking $\rho_u$ correctly is very important for performance. This penalty term typically varies within large range; for example for our experiments it can be set within $[10^{-4} ~,~ 10^2]$. Setting this value correctly is a counter-intuitive process as it is not clear which value would yield good results, and often times may require a large number of cross validations. Another drawback is, $\rho_u$ is fixed from the beginning and remains the same for the entire stream. This may not be desirable as changing the regularization amount at different time steps might improve performance. For these reasons it is beneficial to (i) replace $\rho_u$ with another parameter that is easier to set, and (ii) have time-varying regularization. 

To address both issues we consider the following problem
\begin{align} \label{eq:ft}
	&\min_{\bU_t,\bv_t} \| \bU_t - \bbU \|_F^2 + \| \bv_t - \bbv \|_2^2 \\
	&\text{s.t.}~~ \| \bx_t - \bU_t^\top \bv_t \|_2^2 \leq \eps 
\end{align}

Therefore, instead of $\rho_u$ and $\rho_v$, we introduced $\eps$. This new parameter enforces the matrix factorization to keep the approximation error below $\eps$. Since this error bound is fixed at the beginning, we call this fixed tolerance (FT) matrix factorization. Here $\bbU$ and $\bbv$ are defined the same way in FP. Based on the generative model in Eq. \eqref{eq:gen}, we can interpret the optimization problem of Eq. \eqref{eq:ft} as follows: FT aims finding the point estimates that are closest to the priors, while keeping the likelihood of the observation above a threshold determined by $\eps$.

\subsubsection{Update for $\bU_t$}

The objective function of FT is optimized in an alternative fashion; for a fixed $\bv_t$ the Lagrangian is given by
\begin{align}
	\L(\bU_t,\lam) = \| \bU_t - \bbU \|_F^2 + \lam \| \bx_t - \bU_t^\top \bv_t \|_2^2 - \lam \eps ~,
\end{align}
which yields the following update for $\bU_t$.
\begin{align}
	\bU_t &\gets (\lam^{-1} \bI + \bv_t \bv_t^\top)^{-1} (\lam^{-1} \bbU + \bv_t \bx_t^\top) ~.
\end{align}
Comparing to the update in \eqref{eq:fp_update}, the equivalency is seen by setting $\lam = \rho_u^{-1}$. Since the Lagrange multiplier changes value with every update of $\bv_t$ we now have a varying regularizer, as opposed to the fixed regularizer of FP.

Clearly the main issue with the FT approach is the structure of the constraint set: As the Lagrangian reveals, it is straightforward to compute the gradients, but projection onto the constraint set is not. In fact, this problem is an instance of a quadratically constrained quadratic program (QCQP), for which, there is no closed-form solution in general \cite{Boyd_2004}. When the problem is convex, off-the-shelf solvers can be employed to find the global optimum. To check convexity, let $\sI_t$ denote the index of observed $\bx_t$ entries; dropping the time indices and re-writing the optimization problem gives
\begin{align}
	&\min_{\{\bu_i\}_{i \in \sI_t}} \sum_{i \in \sI_t} \| \bu_i - \bbu_i \|_2^2 ~~\text{s.t.}~~ \sum_{i \in \sI_t} (\bx_i - \bu_i^\top \bv)^2 \leq \eps \nn \\
	&\equiv \min_{\{\bu_i\}_{i \in \sI_t}} \sum_{i \in \sI_t} \bu_i^\top \bI \bu_i - 2\bu_i^\top \bbu_i + \bbu_i^\top \bbu_i \nn \\
	&\quad\quad\quad \text{s.t.} ~~ \sum_{i \in \sI_t} \bu_i^\top \bv \bv^\top \bu_i - 2 x_i \bu_i^\top \bv + x_i^2 ~.
\end{align}
Since both $\bI$ and $\bv \bv^\top$ are positive semidefinite, the problem is indeed convex. Furthermore, the problem is feasible, as the feasible set always contains infinitely many elements (see Appendix \ref{app1} more about this). On the other hand, using a convex solver for every time step is a very inefficient task, and defeats the purpose of having a scalable online algorithm. The best case would be to have a closed form solution to this problem, which does not require any iterative schemes or user-specified learning rates, similar to the FP problem.

It turns out, this goal is indeed achievable for the FT constraint. Define the following constants: \footnote{While $\bv_t$ is a variable of the overall problem \eqref{eq:ft}, it is indeed fixed while updating for $\bU_t$.}
\begin{align}\label{eq:ft_const}
	c_1 = \bv_t^\top \bbU \bx_t ~, c_2 = \|\bv_t\|_2^2 ~, c_3 = \|\bx_t\|_2^2 ~, c_4 = \|\bbU^\top \bv_t\|_2^2 ~,
\end{align}
and set the Lagrange multiplier as
\begin{align} \label{eq:ft_lagm_U}
	\lamst = -\frac{1}{c_2} + \frac{\sqrt{c_3 + c_4 - 2 c_1}}{\sqrt{\eps}c_2} ~.
\end{align}
Then the update for $\bU_t$ is given by
\begin{align}\label{eq:ft_poly_U}
	\bU_t &\gets (\bI + \lamst \bv_t \bv_t^\top)^{-1} (\bbU + \lamst \bv_t \bx_t^\top) ~.
\end{align}   
The derivation for this result turns out to be quite tedious; but we present a condensed version in Appendix \ref{app1} with all key steps.

\subsubsection{Update for $\bv_t$}

Proceeding in a similar manner, for fixed $\bU_t$ the Lagrangian is
\begin{align} \label{eq:v_lagrangian}
	\L(\bv_t,\lam) = \| \bv_t - \bbv \|_2^2 + \lam \| \bx_t - \bU_t^\top \bv_t \|_2^2 - \lam \eps ~,
\end{align}
and the update for $\bv_t$ becomes
\begin{align} \label{eq:ft_v_update}
	\bv_t \gets (\lambda^{-1} \bI + \bU_t \bU_t^\top)^{-1} (\lambda^{-1} \bbv + \bU_t \bx_t) ~.
\end{align}
Once again it is straightforward to verify that this problem is convex. However, a more fundamental issue arises when we take a close look at the constraint. In particular, for a given threshold $\eps$ it is not clear if we can find a $\bv_t$ such that the constraint is satisfied. As an example, when the system of equations is over-determined, the smallest error we can achieve is the least squares error. As $M_t \gg d$, such systems arise frequently during the optimization. When the system is underdetermined and the least squares error is greater than $\eps$, the optimization will terminate as the value of $\bv_t$ at previous iteration will still be the best possible (more details are given in Appendix B). Unfortunately, the feasible set contains many such \emph{isolated} points. Therefore if we seek the minimum norm solution for $\bv_t$ in Eq. \eqref{eq:ft} it is highly likely that the optimization will terminate early on, resulting in poor performance.

To circumvent this we propose the following modification: Instead of finding the minimum-norm solution, we consider updating $\bv_t$ using the Lagrangian in Eq. \eqref{eq:v_lagrangian}, for a \emph{fixed} $\lambda$. In particular, this gives the same update equation with FP, which is given in Eq. \eqref{eq:fp_update}. From the optimization perspective, what we are doing is replacing the least norm update with the ridge regression update (FT vs. FP). The two problems have the same solution when the Lagrange multiplier is the same. Therefore, while the two updates are not equivalent, structurally they are similar to each other. The case when the two updates are the same is established next.

\textbf{Remark: } The FT update for $\bv_t$ and the ridge regression update in Eq. \eqref{eq:ft_v_update} are the same when $\lambda$ is selected as the root of the following polynomial yielding smallest $\| \bv_t \|_2^2$
\begin{align} \label{eq:ft_poly_v}
	p(\lambda) &= \sum_{i=1}^d (-2\rho-\psi_i) c_{1,i}^2 \prod_{j \neq i} (\rho + \psi_j)^2 + \sum_{i=1}^d \rho^2 \psi_i c_{2,i}^2 \prod_{j \neq i} (\rho + \psi_j)^2 \nn \\
	&- 2 \rho^2 \sum_{i=1}^d c_{1,i} c_{2,i} \prod_{j \neq i} (\rho + \psi_j)^2 - (\eps - \bx^\top \bx) \prod_{j=1}^d (\rho + \psi_j)^2 
\end{align}
This result is derived in Appendix \ref{app2}. The final version of FT is summarized in Algorithm \ref{alg:ft}; we can see that both FP and FT are structurally similar, and the key difference is the computation of $\lam$ and the update of $\bU_t$.

\begin{algorithm}[t]
\caption{Fixed Tolerance Matrix Factorization (FT)}\label{alg:ft}
\begin{algorithmic}[1]
\State \textbf{Require: } $\bx_t$, $\sI_t$, $\eps$, $\rho_v$, $\bbU$, $\bbv$, ${\tt max\_ite}$
\State \textbf{Return: $\bU_t$, $\bv_t$}
\vspace{.05in}
\State Re-assign $\bbU \gets \bbU(:,\sI_t)$
\For{$i=1,\dots,{\tt max\_ite}$}
\vspace{.05in}
\State $\bv^{(i)} \gets (\rho_v \bI + \bU^{(i-1)} \bU^{(i-1)\top})^{-1} (\rho_v \bbv + \bU^{(i-1)} \bx_t)$
\State Compute $c_1$, $c_2$, $c_3$, $c_4$ as in \eqref{eq:ft_const}.
\vspace{.05in}
\State $\lamst \gets -\frac{1}{c_2} + \frac{\sqrt{c_3 + c_4 - 2 c_1}}{\sqrt{\eps}c_2}$
\vspace{.05in}
\State $\bU^{(i)} \gets (\bI + \lamst \bv^{(i)} \bv^{(i)\top})^{-1} (\bbU + \lamst \bv^{(i)} \bx_t^\top)$
\EndFor
\State Update $\bU_t(:,\sI_t) \gets \bU$ and $\bU_t(:,\sI_t^c) \gets \bU_{t-1}(:,\sI_t^c)$.
\State Update $\bv_t \gets \bv$.
\end{algorithmic} 
\end{algorithm}

\subsection{Zero Tolerance Constraint}
\label{sec:zt}

So far, two different approaches to online matrix factorization have been developed; the key difference between FP and FT is that, while FP fixes a penalty parameter $\rho_u$, FT fixes the error tolerance $\eps$. While fixing one parameter makes the other vary across iterations, from the user perspective, what we have done is equivalent to replacing one \emph{tunable} parameter with another. A key question is then, if we can make this factorization \emph{parameter free}. This motivates the special case $\eps=0$ which we refer to as zero tolerance (ZT) matrix factorization. Interpreting from the perspective of the generative model in Eq. \eqref{eq:gen}, ZT estimates the latent factors $\bU_t$ and $\bv_t$ that are as close to the prior as possible, while making zero approximation error on $\bx_t$. This corresponds to setting the noise term $\bna{\bx}{t} = \b0$, i.e. the measurements are the ground truth and the latent factors should perfectly reconstruct them.

The optimization problem reduces from Eq. \eqref{eq:ft} as
\begin{align} \label{eq:zt}
	&\min_{\bU_t,\bv_t} \| \bU_t - \bbU \|_F^2 + \| \bv_t - \bbv \|_2^2 \\
	&\text{s.t.}~~ \bU_t^\top \bv_t = \bx_t ~. 
\end{align}
From this, it is immediately seen that the formulation is equivalent to that of nuclear norm minimization problems \cite{Recht_2009, Candes_2010}. The two changes made are, we consider the factored form of the nuclear norm \cite{Recht_2010} and we do online optimization, as opposed to batch.

Now consider optimizing $\bU_t$ while $\bv_t$ is fixed. As discussed before, the linear system $\bU_t \bv_t = \bx_t$ is underdetermined for a variable $\bU_t$. Eq. \eqref{eq:zt} suggests finding the solution with the least Frobenius norm. This generalizes the least norm problem that is considered for linear underdetermined systems to the matrix case. As the system is underdetermined, the feasible set will contain infinitely many points; this follows the same argument we use in Appendix \ref{app1}. Then $\bU_t$ can be found with Lagrange multipliers. After an inconsequential rescaling, the Lagrangian is given by
\begin{align}
	\L(\bU_t,\blam) = \frac{1}{2} \| \bU_t - \bU_{t-1} \|_F^2 + \blam^\top (\bx_t - \bU_t \bv_t) ~. 
\end{align}
Here note the boldface notation for $\blam$, since in this case there are $M_t$ equality constraints. The stationarity conditions are
\begin{alignat}{2}
	\nabla_{\bU_t} \L(\bU_t,\blam) &= 0 ~~ &&= \bU_t - \bU_{t-1} + \bv_t \blam^\top ~,  \nn \\
	\nabla_{\blam} \L(\bU_t,\blam) &= 0 ~~ &&= \bU_{t-1}^\top \bv_t - \bx_t ~.
\end{alignat}
The solution is then given by
\begin{equation}
	\blam = \frac{\bU_{t-1}^\top \bv_t - \bx_t}{\bv_t^\top \bv_t} ~~,~~ \bU_t = \bU_{t-1} - \bv_t \blam^\top ~.
\end{equation}
Note that while $\blam$ is changing over time, it can no longer be seen as the inverse regularizer of the FP term; because $\eps$ is no longer a tunable parameter, but hard-coded to zero. This is advantageous in that, the user does not have to cross-validate to find a good value for it. On the other hand, as we will show in the experiments, $\eps=0$ requirement can become too restrictive in some cases, in which case it will require a higher value of $d$ - the factorization rank.

On the other hand, the update for $\bv_t$ suffers from the same problem we discussed in the previous section. For the ZT constraint, consider the case when $\bU_t^\top \bv_t = \bx_t$ is overdetermined for variable $\bv_t$. Then the smallest error achievable will be given by the least squares solution, which satisfies $\eps_\text{ls} > 0$, so the feasible set is empty. However, since the the optimization is done in an alternating manner, this worst case does not occur in practice. In particular, at iteration $i-1$ we have already found $(\bU_{t-1}^{(i-1)},\bv_{t-1}^{(i-1)})$ such that $\bU_{t-1}^{(i-1)\top} \bv_{t-1}^{(i-1)} = \bx_t$. This means, when we update for $\bv_t^{(i)}$ for a fixed $\bU_t^{(i-1)}$, the feasible set will contain at least $\bv_t^{(i-1)}$. The problem, however, is that, if this is the only point contained in the feasible set, the optimization will terminate, and once again there is a risk of getting stuck early in the process. Once again this can be circumvented by replacing the least norm solution with the $\ell_2$ regularized one, for which we re-introduce $\rho_v$ as a small set-and-forget parameter and update
\begin{align}
	\bv_t \gets (\rho_v \bI + \bU_t \bU_t^\top)^{-1} (\rho_v \bbv + \bU_t \bx_t) ~.
\end{align}
In summary, ZT is simply the special case of FT where we set $\eps=0$. As $\rho_v$ is a small constant, ZT can be treated as a parameter-free matrix factorization method; which is advantageous for large and/or streaming data as tuning/cross-validation would be difficult. ZT is summarized in Algorithm \ref{alg:zt}.

\begin{algorithm}[t]
\caption{Zero Tolerance Matrix Factorization (ZT)}\label{alg:zt}
\begin{algorithmic}[1]
\State \textbf{Require: } $\bx_t$, $\sI_t$, $\rho_v$, $\bbU$, $\bbv$, ${\tt max\_ite}$
\State \textbf{Return: $\bU_t$, $\bv_t$}
\vspace{.05in}
\State Re-assign $\bbU \gets \bbU(:,\sI_t)$\For{$i=1,\dots,{\tt max\_ite}$}
\vspace{.05in}
\State $\bv^{(i)} \gets (\rho_v \bI + \bU^{(i-1)} \bU^{(i-1)\top})^{-1} (\bU^{(i-1)} \bx_t)$
\State $\blam \gets (\bbU \bv^{(i)} - \bx_t) / (\bv^{(i)\top} \bv^{(i)})$
\vspace{.05in}
\State $\bU^{(i)} \gets \bbU - \bv^{(i)} \blam^\top$
\EndFor
\State Update $\bU_t(:,\sI_t) \gets \bU$ and $\bU_t(:,\sI_t^c) \gets \bU_{t-1}(:,\sI_t^c)$.
\State Update $\bv_t \gets \bv$.
\end{algorithmic} 
\end{algorithm}

\section{Optimum Sequence Prediction}
\label{sec:pred}

\begin{algorithm}[t]
\caption{Online Forecasting Matrix Factorization}\label{alg:ofmf}
\begin{algorithmic}[1]
\State \textbf{Require: } $\bX$, $\sI$, $d$, $r_0$, $\rho_u$ (FP), $\eps$ (FT), $\rho_v$, ${\tt max\_ite}$
\State \textbf{Return: } $\forall t$: $\hbx_t$, $\bU_t$, $\bv_t$, $\btheta_t$
\vspace{.05in}
\State Initialize $\bU^{(0)} \gets {\tt rand}(M,d)$, $\bv^{(0)} \gets {\tt rand}(d,1)$. 
\State Initialize $\br_{l,0} \gets r_0 \bI$, $\br_{r,0} \gets 0$.
\For{$t=1,\ldots,T$}
	\State {\tt // Forecast Step}
	\vspace{.01in}
	\State $\bbU \gets \bU_{t-1}$, $\bbv \gets \sum_{l=1}^P \theta_l \bv_{t-l}$ \textsuperscript{$\dagger$}
	\vspace{.02in}
	\State Forecast: $\hbx_t = \bbU^\top \bbv$
	\State {\tt // E-Step} 
	\State Use one of the following:\\
	\begin{itemize}[leftmargin=1cm]
	\item ${\tt FP}(\bX_t,~ \sI_t, ~\rho_u, ~\rho_v, ~\bbU, ~\bbv, ~{\tt max\_ite})$
	\item ${\tt FT}(\bX_t,~ \sI_t, ~\eps, ~\rho_v, ~\bbU, ~\bbv, ~{\tt max\_ite})$
	\item ${\tt ZT}(\bX_t,~ \sI_t, ~\rho_u, ~\bbU, ~\bbv, ~{\tt max\_ite})$
	\end{itemize}
	\vspace{.05in}
	\State {\tt // M-Step} 
	\If{$t>P$}
		\State $\bP_t \gets [\bv_{t-1},\ldots,\bv_{t-P}]$
		\vspace{.02in}
		\State $\br_{l,t} \gets \br_{l,t-1} + \bP_t^\top \bP_t$
		\vspace{.02in}
		\State $\br_{r,t} \gets \br_{r,t-1} + \bP_t^\top \bv_t$
		\vspace{.02in}
		\State $\btheta_t \gets \br_{l,t}^{-1} \br_{r,t}$
	\EndIf	
\EndFor
\vspace{.1in}
\State \textsuperscript{$\dagger$} Corner case: If $t=1$ set $\bbU = \b0$ and $\bbv = \b0$. If $1<t<P$ set $\bbv = \bv_{t-1}$. Otherwise use the update in Line 7.
\end{algorithmic} 
\end{algorithm}

Section \ref{sec:mf} detailed three matix factorization approaches with smoothness penalty on the matrix $\bU$, which contains the coefficients required to construct the $M \times T$ matrix $\bX$ from the $d \times T$ matrix $\bV$, or equivalently the set $[\bv_T]$. As discussed in Section \ref{sec:background}, each column $\bv_t$ corresponds to an observation of $d$ \emph{compressed} time series. The algorithms presented in the previous section generate the $[\bv_T]$ sequence in an online manner. 

When the columns of the original time series matrix $\bX$ have correlation, it is natural to expect a similar structure in $\bV$. For this reason, similar Equation \ref{eq:ar_model} we define an AR model for the columns of $\bV$ as:
\begin{align}\label{eq:ar_latent}
	\bv_t = \theta_1 \bv_{t-1} + \ldots + \theta_P \bv_{t-P} + \bna{\bv}{t} ~.
\end{align}   
The task is, then, to find the coefficient vector $\btheta$ given the vector observations $[\bv_T]$. Similar to the matrix factorization problem, there is no single answer here; yet it is desirable to have an estimation scheme with flexibility and minimal assumptions. For these purposes, the LMMSE estimator turns out to be a good choice, as (i) it only needs assumptions about the first and second order statistics of the random variables involved, and (ii) optimizing the mean square error is numerically stable, as opposed to implementing the BLUE estimator.

Before going any further, we introduce some additional terminology. First note that, \eqref{eq:ar_latent} corresponds to $\bv_t = \bP_t \btheta$ where $\bP_t = [\bv_{t-1} ~ \cdots ~ \bv_{t-P}]$ is a $d \times P$ \emph{patch} matrix of the previous $P$ columns. The entire collection of such matrices are obtained by \emph{vertically} stacking the patch matrices as $\bP = [\bP_1^\top ~ \cdots ~ \bP_T^\top]^\top$ which is a $Td \times P$ matrix. On the other hand, stacking the observation vectors vertically, we obtain $\bp = [\bv_1^\top ~ \cdots ~ \bv_T^\top]^\top$, a vector with $Td$ elements. The vector $\bn$ is defined similarly. Here we note an important detail: At the beginning of Section \ref{sec:background} we assumed that the observations start at $T=1$. This means, to obtain a patch matrix which does not contain any zero-column, we should start constructing matrices $\bP$ and $\bp$ from the index $t=P+1$. Throughout this section, however, we omit this detail to avoid clutter in the equations. The final algorithm we present, however, covers this corner case.

Now for the entire set of $[\bv_T]$, we have the relation
\begin{equation}
	\bP \btheta + \bn = \bp ~,
\end{equation}
which means each vector observation contains information about a latent vector $\btheta$. Note that this setting is different from Kalman Filter \cite{Kalman_1960}, where the vector $\btheta$ itself is a time-varying latent variable. A good estimator should provide accurate values for $\btheta$, with minimal assumptions about the distributions of the random variables involved. To that aim, let the noise distribution have the following first- and second-order statistics
\begin{equation}
	\E[\bn_t] = \b0 ~,~ \E[\bn_{t_1} \bn_{t_2}^\top] = \bSigma_{\bn} ~  \delta(t_1,t_2) ~,
\end{equation}
where $\delta(t_1,t_2)$ is the Kronecker delta function. This is a white noise process with stationary covariance; but no assumptions about its p.d.f. has been made. For the parameter $\btheta$ there are two options: (i) it can be treated as an unknown deterministic parameter (classical inference) or (ii) it can be modeled as a random variable (Bayesian inference). For reasons that will be clear shortly, we choose the second route, and assume that $\btheta$ satisfies the following
\begin{equation}
	\E[\btheta] = \b0 ~,~ \E[\btheta \btheta^\top] = \bSigma_{\btheta} ~.
\end{equation} 
We also note that $\btheta$ and $\bn$ are assumed independent. Given all the ingredients, we restrict ourselves to the linear estimators of form $\hbtheta = \bW \bp$; to conform the equality, the weight matrix $\bW$ must be $dT \times P$. We now want to minimize the mean square error as
\begin{equation}
	\text{MSE} = \min_{\hbtheta} \| \btheta - \hbtheta \|_2^2 \equiv \min_{\bW} \| \btheta - \bW \bp \|_2^2 ~.
\end{equation}
We can write
\begin{align}
	\text{MSE}(\bW) &= \E ~ [(\btheta - \bW \bp)^\top (\btheta - \bW \bp)] \nn \\
	&= \E ~ \tr ~  [(\btheta - \bW \bp) (\btheta - \bW \bp)^\top] \nn \\
	&= \tr ~ \E ~ [(\btheta - \bW \bp) (\btheta - \bW \bp)^\top] \nn \\
	&= \tr ~ \{ \bSigma_{\btheta} + \bW \bP \bSigma_{\btheta} \bP^\top \bW^\top  + \bW \bSigma_{\bn} \bW^\top \nn \\
	&\quad\quad\quad - \bSigma_{\btheta} \bP^\top \bW^\top - \bW \bP \bSigma_{\btheta} \} ~,
\end{align}
which reveals that the optimum estimator can be found by the stationarity condition $\nabla_{\bW} \text{MSE}(\bW) = 0$. Taking the matrix derivatives with respect to trace, the optimum solution becomes
\begin{equation}\label{eq:weig1}
	\bW = \bSigma_{\btheta} \bP^\top [\bSigma_{\bn} + \bP \bSigma_{\btheta} \bP^\top]^{-1}.
\end{equation}
Matrix inversion lemma asserts, for conformable matrices $\bM_1$, $\bM_2$, $\bM_3$, $\bM_4$
\begin{align}\label{eq:MIL}
	[\bM_1 + \bM_2 \bM_3& \bM_4]^{-1} = \bM_1^{-1} - \bM_1^{-1} \bM_2 \times \nn \\
	&[\bM_3^{-1} + \bM_4 \bM_1^{-1} \bM_2]^{-1} \bM_4 \bM_1^{-1}
\end{align}
given the inverses exist.

Then \eqref{eq:weig1} can be re-written as
\begin{align} \label{eq:weig2}
	\bW &= \bSigma_{\btheta} \bP^\top \[ \bSigma_{\bn}^{-1} - \bSigma_{\bn}^{-1} \bP [\bSigma_{\btheta}^{-1} + \bP^\top \bSigma_{\bn} \bP]^{-1} \bP^\top \bSigma_{\bn}^{-1} \] \nn \\
	&= [\bP^\top \bSigma_{\bn}^{-1} \bP + \bSigma_{\btheta}^{-1}]^{-1} \bP^\top \bSigma_{\bn}^{-1} ~.
\end{align}
Here the second line can be obtained by a straightforward algebraic manipulation. The LMMSE estimator is then given by
\begin{equation} \label{eq:that_full}
	\hbtheta = [\bP^\top \bSigma_{\bn}^{-1} \bP + \bSigma_{\btheta}^{-1}]^{-1} \bP^\top \bSigma_{\bn}^{-1} \bp ~. 
\end{equation}
It immediately follows from this functional form, that LMMSE estimator reduces to BLUE when a non-informative prior is chosen. This can be written -with a slight abuse of notation- as $\bSigma_{\btheta} = \infty \bI$. In addition, when $\bSigma_{\bn} = \bI$, BLUE coincides with the ordinary least squares estimator, yielding Gauss-Markov theorem \cite{Hastie_2009}. For this paper we consider the case $\bSigma_{\bn} = \bI$; however $\bSigma_{\btheta} = r_0 \bI$ for a tunable parameter $r_0$.

Moving from \eqref{eq:weig1} to \eqref{eq:weig2} reveals an important structure. Set $\bSigma_{\bn} = \bI$. Using matrix partitioning properties \cite{Gentle_2007}, \eqref{eq:that_full} can be written as
\begin{equation}
	\hbtheta = \[ \sum_{t=1}^T \bP_t^\top \bP_t + \bSigma_{\btheta}^{-1} \]^{-1} \[ \sum_{t=1}^T \bP_t^\top \bv_t \] ~.
\end{equation}
This suggests we can compute the two terms on the left- and right- hand side \emph{recursively}. In particular, define $\br_{l,0} = \bSigma_{\btheta}^{-1}$ and $\br_{r,0} = 0$ and the recursions
\begin{equation}
	\br_{l,t} = \br_{l,t-1} + \bP_t^\top \bP_t ~,~ \br_{r,t} = \br_{r,t-1} + \bP_t^\top \bv_t ~.
\end{equation}
Then, at any given time $t$, we can obtain a most up-to-date weight estimate simply by $\hbtheta_t = \br_{l,t}^{-1} \br_{r,t}$. We now have a fully online algorithm for both factorizing the incoming data matrix, and estimating the AR coefficients for the compressed time series.

We summarize the entire algorithm in Algorithm \ref{alg:ofmf}.

\begin{figure*}[t]
\centering
\subfloat[]{\includegraphics[width=2.5in]{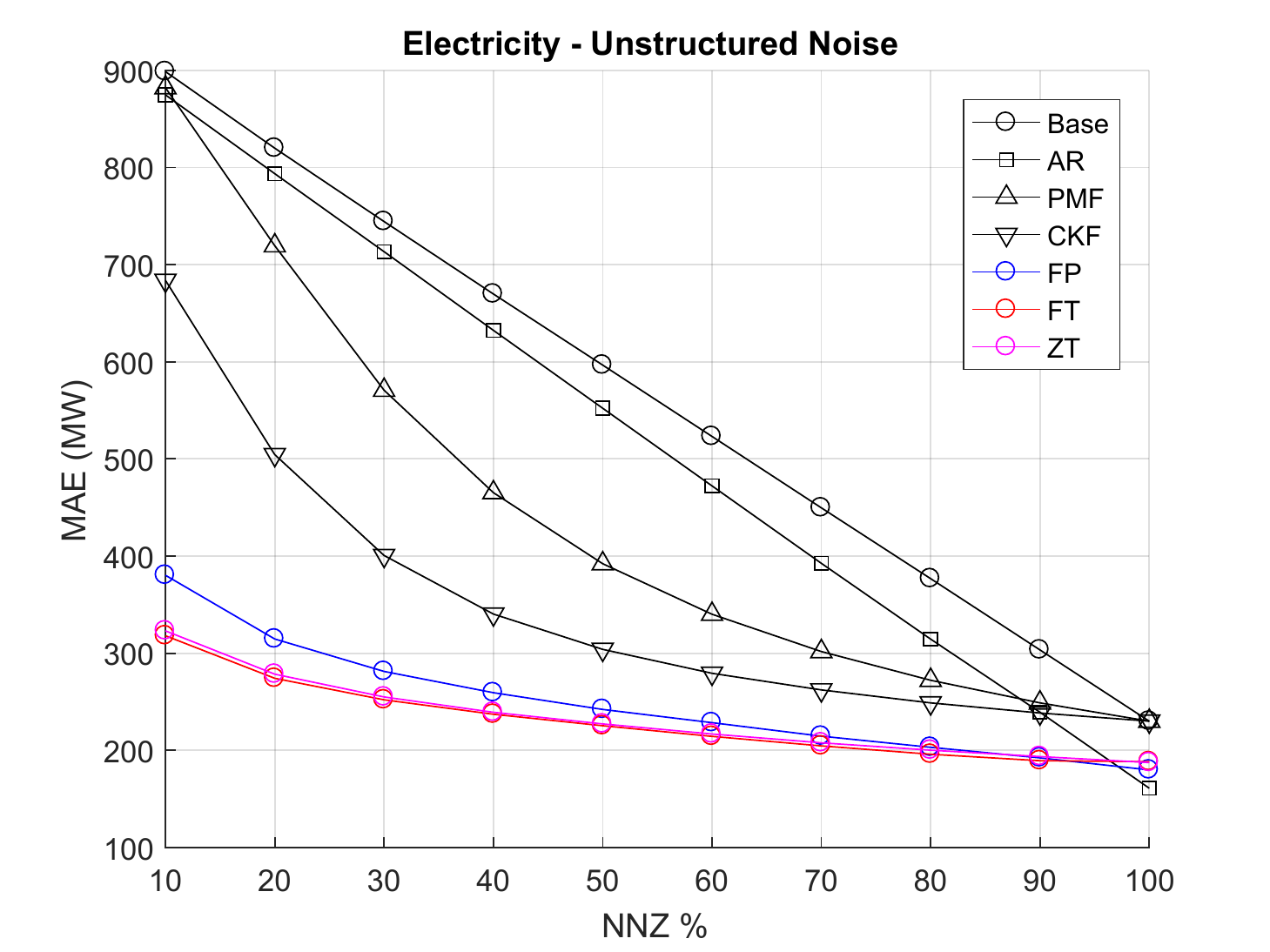}\label{fig:elec_mae_1}}
\hfil
\subfloat[]{\includegraphics[width=2.5in]{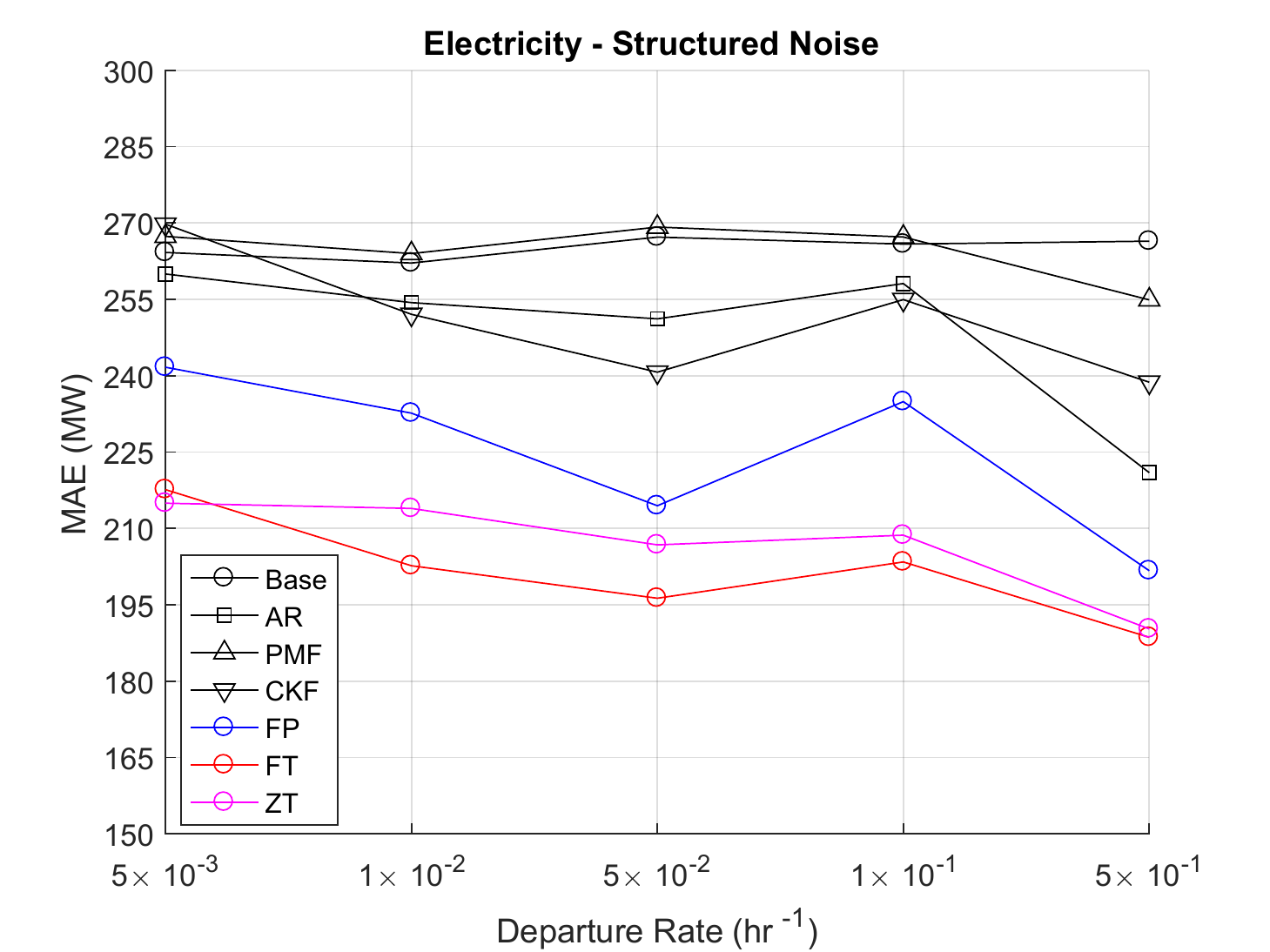}\label{fig:elec_mae_2}}
\caption{Performance comparison of seven predictors listed in the beginning of this section, for the electricity dataset. (a) The sparsity pattern is unstructured, and 20 sets of experiments are performed for 10 different levels. (b) The sparsity pattern is structured, and 20 sets of experiments are performed for 5 different departure rates.}
\label{fig:elec_mae}
\end{figure*}

\section{Experiments}
\label{sec:exp}

In this section we test our proposed methodology using two time-series datasets downloaded from UCI machine learning repository. These datasets are:
\begin{itemize}
	\item Electricity:\footnote{https://archive.ics.uci.edu/ml/datasets/ElectricityLoadDiagrams20112014} Hourly power consumptions (Mega Watts) of 370 customers, recorded between Jan 1, 2012 to Jan 1, 2015, in Portugal. This gives a matrix of 370 rows and 26,304 columns, with 9,732,480 entries.
	\item Traffic:\footnote{ https://archive.ics.uci.edu/ml/datasets/PEMS- SF} Hourly occupancy rates of 963 roads in Bay Area, California, recorded between Jan 1, 2008 and Mar 30, 2009. This time the matrix has 963 rows and 10,560 columns, yielding 10,169,280 entries.
\end{itemize}

For both datasets there are no missing values; this is useful as we can obtain sparse versions at various levels by simply removing elements. For both datasets we experiment with two types of sparsity: (i) Unstructured sparsity where at each time step the corresponding column of $\bX$ is uniformly subsampled. (ii) Structured sparsity where the sparsity follows a geometric process with certain arrival/departure rates. The task of prediction is to predict the observed entries at given time step.

We implement and compare the following:
\begin{itemize}
	\item Base: This is a base estimator, which estimates the current value as the last observation. If the observation at previous time is missing, then it predicts the average of the last observed vector. The base algorithm is important in that, any acceptable forecasting algorithm should perform better than it, as the comparisons would not make sense otherwise. Therefore, it acts as an elementary solution to the forecasting problem as well as a sanity check.
	\item AR($P$): This is simply the AR model of \eqref{eq:ar_model}, implemented on the vector observations. We learn the model in an online manner using the LMMSE estimator derived in Section \ref{sec:pred}. This makes the comparisons with our algorithms fair, as both of them use the same machinery, but on different vector sequences.
	\item PMF: Probabilistic matrix factorization algorithm \cite{Salakh_2008a}. To extend PMF to the online setting, the FP cost function in Eq. \eqref{eq:fp} is used, but of course no AR structure is imposed. 
	\item CKF: Collaborative Kalman Filter \cite{Gultekin_2014}. This is an online approach to matrix factorization problem, where the latent states follow a brownian motion. Unlike PMF, here a full posterior on latent variables can be calculated. 
	\item Naive MF: This is not a competitive algorithm; it corresponds to the model in \eqref{eq:naive_mf}. Here we use this algorithm to show that, it is important to generate the sequence $[\bv_T]$ in a careful manner, and a constraint that does not impose any temporal structure is not suitable for this task. 
	\item FP-MF: Fixed penalty matrix factorization from Section \ref{sec:fp}.
	\item FT-MF: Fixed tolerance matrix factorization from Section \ref{sec:ft}.
	\item ZT-MF: Zero tolerance matrix factorization from Section \ref{sec:zt}.
\end{itemize}

\subsection{Results on Electricity Data} 

For this set of experiments, the tunable parameters of all algorithms considered are set as follows:
\begin{itemize}
	\item AR: $P=24$, $r_0 = 1$
	\item PMF: $d=5$, $\rho_u = 1$, $\rho_v = 10^{-4}$
	\item CKF:\footnote{$\nu_d$ and $\nu_x$ are the drift and measurement noise variance respectively. See \cite{Gultekin_2014} for more details.} $d=5$, $\nu_d = 10^{-4}$, $\nu_x = 10^{-4}$ 
	\item Naive MF: $d=5$, $\rho_u = 1$, $\rho_v = 10^{-4}$
	\item FP-MF: $d=5$, $\rho_u = 1$, $\rho_v = 10^{-4}$, $P=24$, $r_0 = 1$
	\item FT-MF: $d=5$, $\eps = 5 \times 10^{-2}$, $\rho_v = 10^{-4}$, $P=24$, $r_0 = 1$
	\item LN-MF: $d=5$, $\rho_v = 10^{-4}$, $P=24$, $r_0 = 1$
	\item ${\tt max\_ite} = 15$ for all algorithms.
\end{itemize}
These values are found by cross-validation; note that the parameters shared across different algorithms have the same value, which is intuitively satisfying.

The first result we show is the one-step ahead predictive accuracy of seven methods, based on \emph{unstructured} sparsity pattern. This simply means, the entries for each column of $\bX$ are selected/dropped uniformly at random. Unstructured sparsity is important in that, it makes the learning environment adversarial, and the algorithms which are unfit for missing values are strongly affected. For our experiments we use 10 different sparsity levels; in particular, the percentage of observed entries vary from 10\% to 100\% in 10\% increments. This percentage is abbreviated as number of non-zeros (NNZ), based on a boolean observation matrix with 1/0 corresponding to observation/missing value. For each NNZ level we assess the performance using mean absolute error(MAE) which has measure of mega Watts. We do this for all competing predictors, averaging over 20 patterns to ensure statistical significance. In Figure \ref{fig:elec_mae_1} we show the one-step ahead prediction performance of all competing algorithms. When there are no missing observations (NNZ = 100\%) the AR model has the best performance. This is an expected result, as in this case, compressing the dimension of time series from 370 to 5 results in loss of information. As NNZ decreases, however, the optimality of AR quickly disappears, as the three proposed algorithms give the best prediction. In fact, even when NNZ = 90\%, AR still has worse performance. As the sparsity increases, the base predictor and AR suffer the most. On the other hand, PMF and CKF are performing better, as they utilize low-rank representations; in addition, the predictions given by CKF are better overall. Finally, FP/FT/ZT utilize both low rank and temporal regularization, yielding best results. Compared to the others, their prediction suffers significantly less as a function of increasing sparsity. Also, FT and ZT are performing better compared to FP, showing that adaptive regularization is indeed useful.

\begin{figure*}[t]
\centering
\subfloat[]{\includegraphics[width=2.5in]{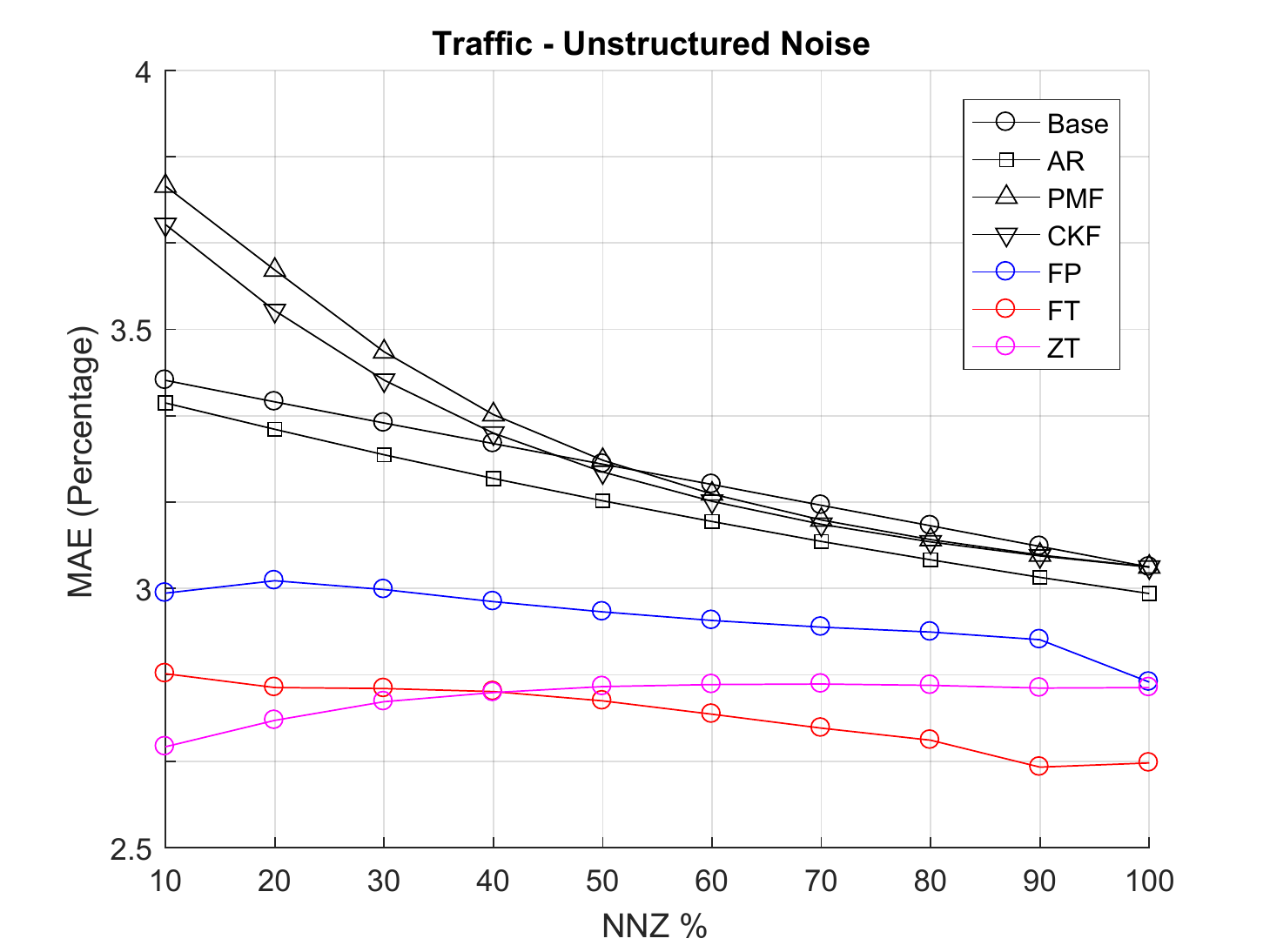}\label{fig:traf_mae_1}}
\hfil
\subfloat[]{\includegraphics[width=2.5in]{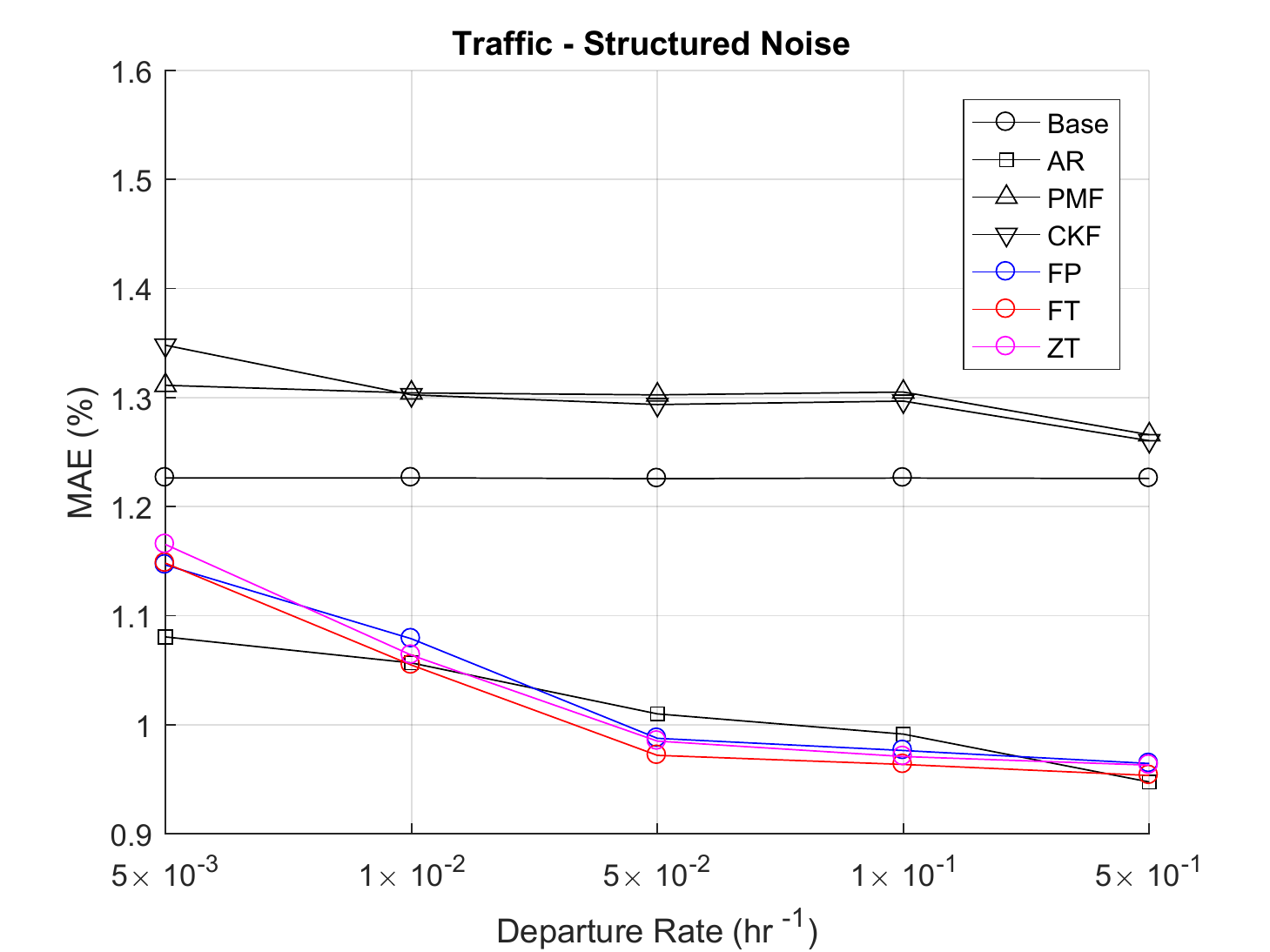}\label{fig:traf_mae_2}}
\caption{Performance comparison of seven predictors listed in the beginning of this section, for the traffic dataset. (a) The sparsity pattern is unstructured, and 20 sets of experiments are performed for 10 different levels. (b) The sparsity pattern is structured, and 20 sets of experiments are performed for 5 different departure rates.}
\label{fig:elec_mae}
\end{figure*}

We now turn to structured sparsity, which is not as adversarial as the previous one. In this case, the missing values corresponds to the arrivals of a random process. For these experiments we use a geometric distribution to find arrival/departure points. This sparsity pattern represents sensor failures or down times; in particular, which is the time between an arrival and departure. A higher arrival indicates increased susceptibility to failure. Note that the rate is simply the success chance of the geometric distribution, and it is inversely proportional to the expected time until arrival/departure. For the electricity data we set the arrival rate to $5 \times 10^{-2}$ and departure takes values in $\{ 5 \times 10^{-3}, 1 \times 10^{-2}, 5 \times 10^{-2}, 1 \times 10^{-1}, 5 \times 10^{-1} \}$. A higher departure rate means lower sparsity. In Figure \ref{fig:elec_mae_2} the prediction MAE is shown as a function of departure rates. The immediate observation is that, even if the sparsity is high (92\% when departure rate is $5 \times 10^{-3}$) none of the algorithms deteriorate as much as they do in Figure \ref{fig:elec_mae_1}. Overall, once again, FT has the best performance, and the margin between FT, ZT and FP is more pronounced. On the other hand, the impute-predict scheme of the baseline predictor as well as the AR predictor also produce acceptable results, while CKF is still better.

\begin{figure}[h]
	\centering
	\includegraphics[scale = .5]{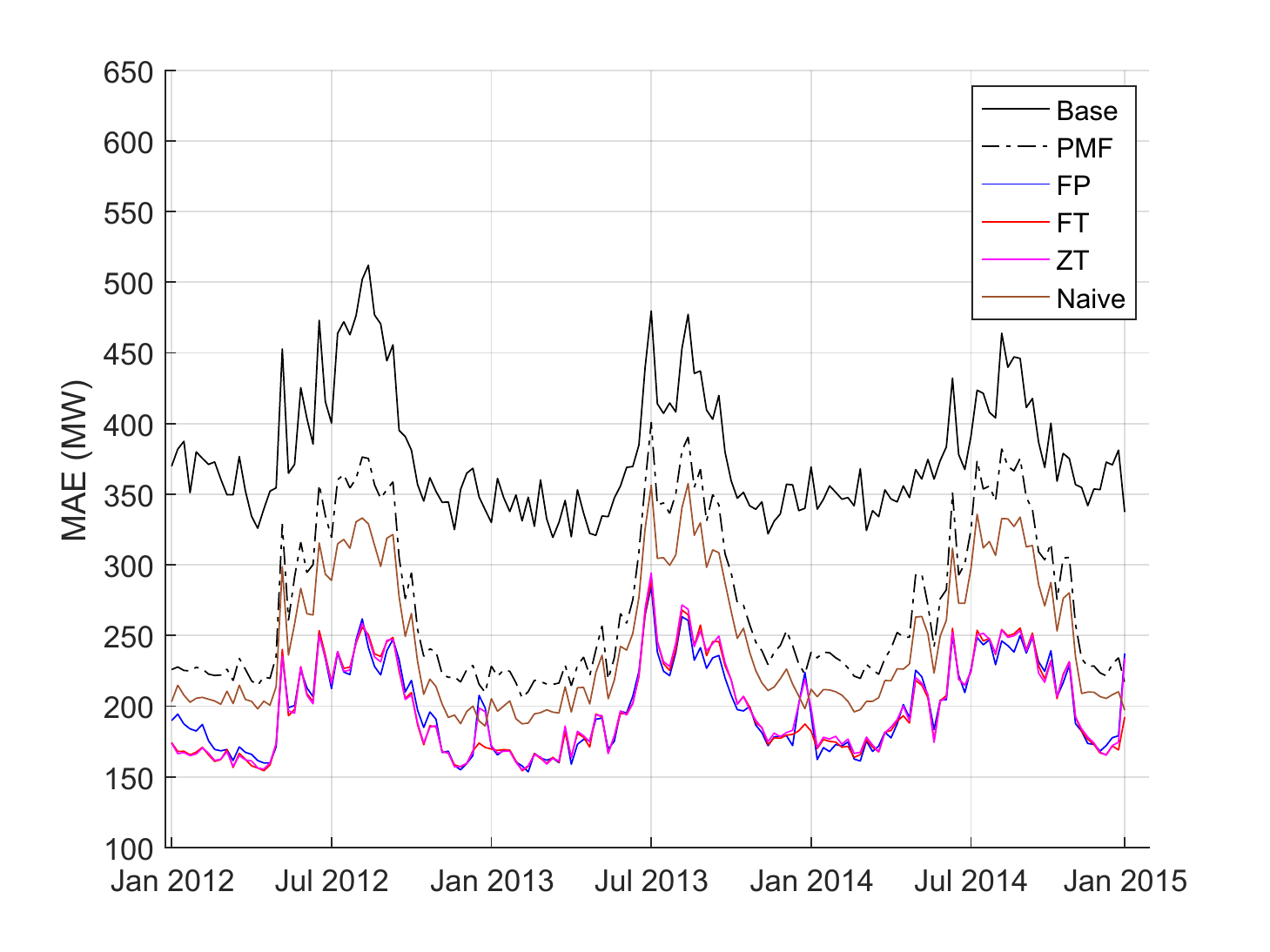}
	\caption{Comparison of PMF, naive MF, FP, FT, and ZT for the Electricity data with unstructured sparsity and NNZ = 80\%. While naive MF can forecast better than PMF, it is inferior to FP/FT/ZT. This shows the importance of generating $[\bv_T]$ properly.}
	\label{fig:elec_naive}
\end{figure}

In the beginning of Section \ref{sec:mf} we have mentioned that rotation and scaling of factor matrices have an important impact on performance. The main argument is, using the projected value of $\bbU$ in regularization, a dependency between the current $\bU_t$ and previous $\bU_{t-1}$ is established, which encourages smooth variation. The alternative regularization in Eq. \eqref{eq:naive_mf} does not have this feature, as the penalty term on $\bU_t$ is centered around zero matrix. Since this constraint does not encourage smoothness, it is expected to do worse compared to the proposed factorizations. We provide evidence for this in Figure \ref{fig:elec_naive}; here the naive approach is compared to FP/FT/ZT as well as PMF. It can be seen that, while the AR model still lets the naive factorization to forecast better than PMF, it is clearly inferior to our methods. This plot also gives more information about the electricity data itself. In particular, observe that all algorithms have higher forecast error during summer times, which indicates that the electric usage during this season is harder to predict in advance. This might be due to increased usage of AC units, which unlike heaters, do not have a regular on/off schedule as well as customers spending more time outside \footnote{This data is collected in Portugal.}.

\begin{figure}[h]
\centering
\subfloat[]{\includegraphics[width = .5\columnwidth]{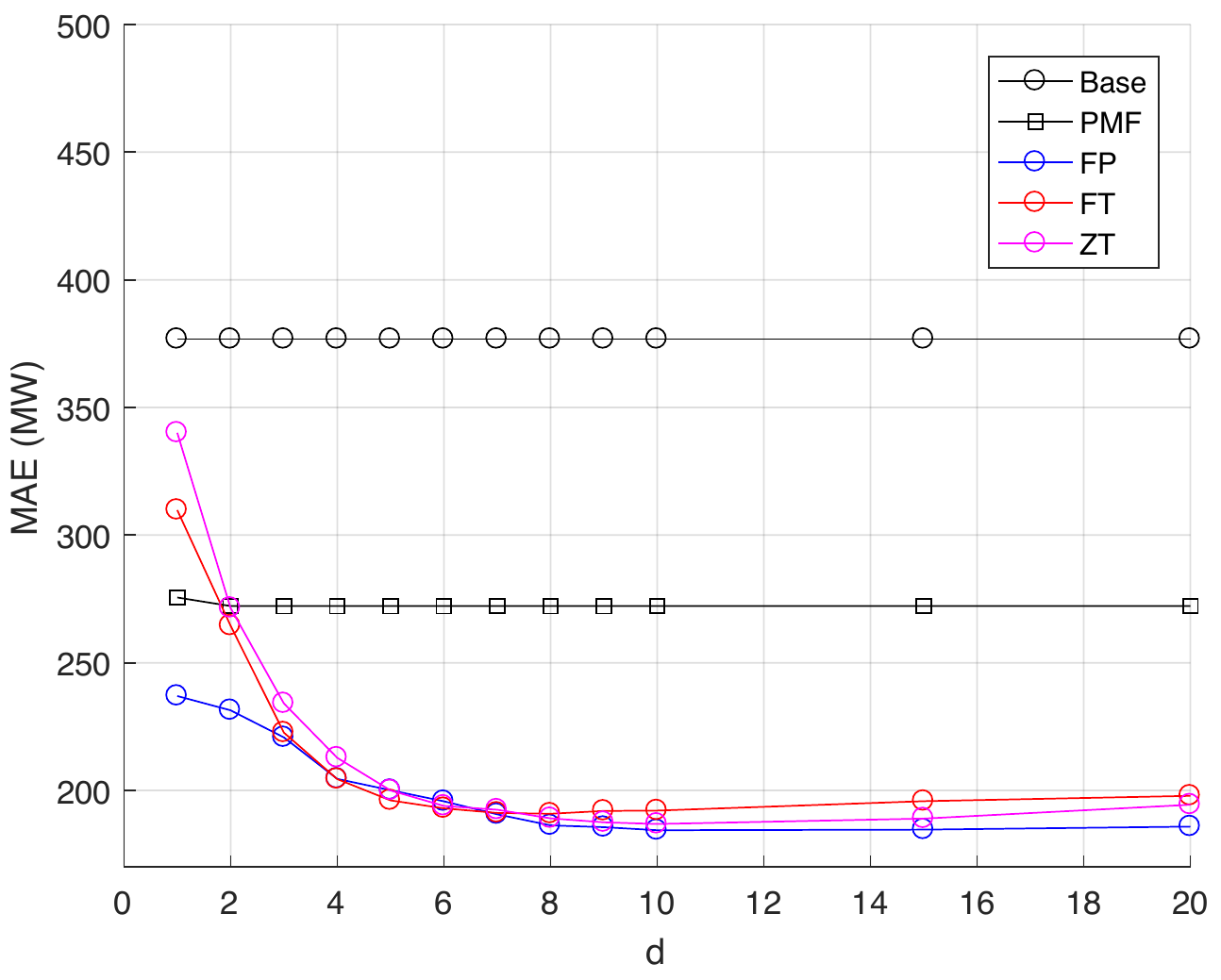}\label{fig:elec_ch_dim}}
\hfil
\subfloat[]{\includegraphics[width = .5\columnwidth]{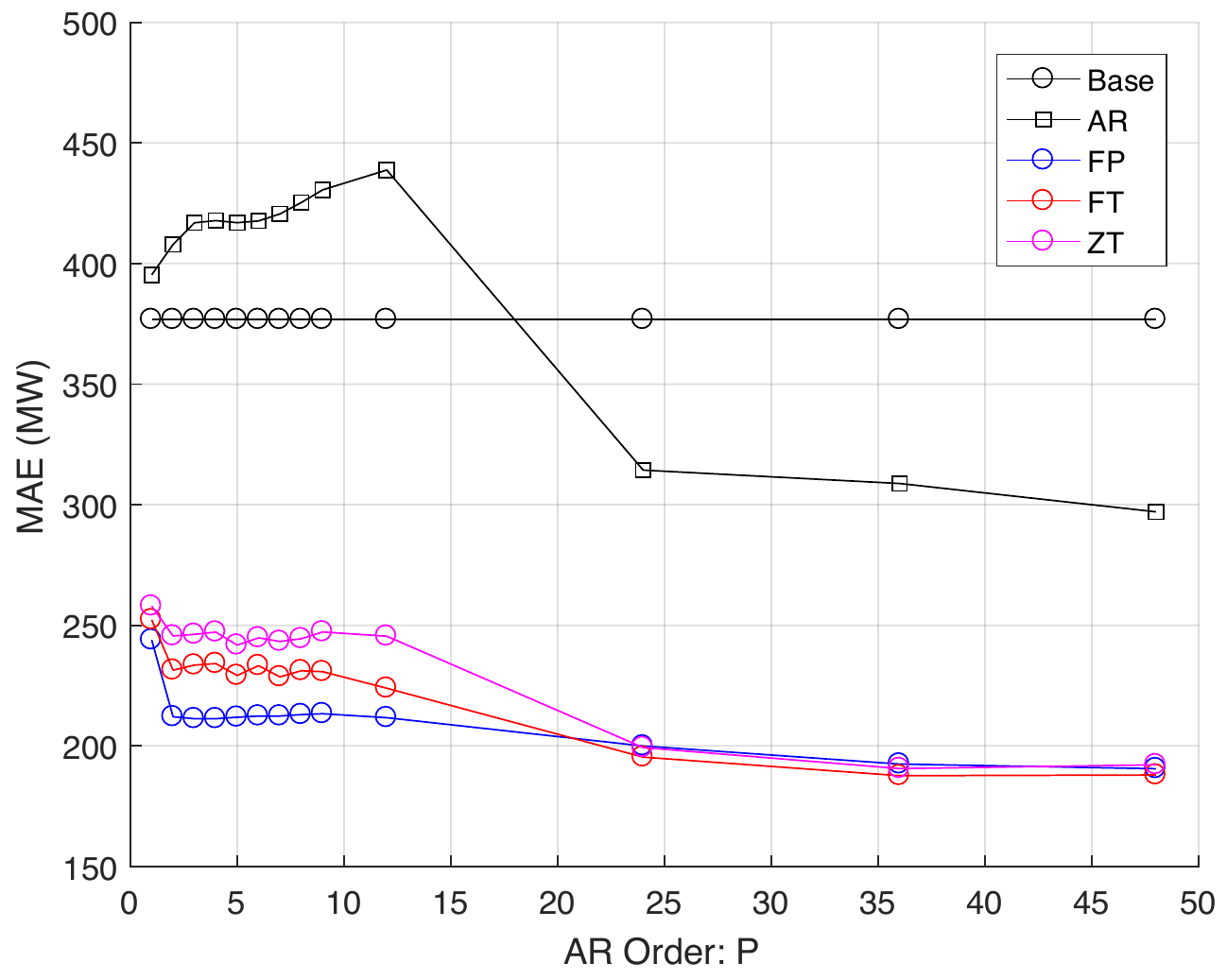}\label{fig:elec_ch_nord}}
\caption{Plot of prediction performances as a function of (a) rank and (b) AR order. Both plots obtained for electricity dataset with unstructured sparsity and NNZ = 80\%.}
\label{fig:elec_ch}
\end{figure}

Dimensionality and AR order are the two most important parameters which determine how the matrix factorization forecasting scheme operates. We next examine performance as a function of these two in Figure \ref{fig:elec_ch}. In panel (a) we plot the performance as a function of latent dimensionality for unstructured noise with 80\% observed entries. Here we show results for PMF as well as FP, FT, and ZT. First note that a dimension of one is giving worst results for all. This is an important sanity check, as it shows the optimum rank is indeed greater than one, and matrix factorization is a suitable choice. The choice $d=10$ produces best results, although we have used $d=5$ for our other experiments, which still produce reliable results with the added benefit of higher compression. Interestingly, when the dimensionality is set low, both FT and ZT perform worse, even worse than PMF. This is because, as dimension decreases the fixed or zero tolerance constraint becomes more restrictive and compromises performance. The compromise for ZT, in turn, is greater than FT as expected, as it is a zero training error constraint. Therefore, when $d$ needs to be low, we can use FT instead of ZT with $\eps>0$, which provides slackness for better factorization. In panel (b) we show results as a function of AR order where $P \in \{1,2,3,4,5,6,7,8,9,12,24,36,48\}$. Note that there is a jump in performance as P moves from 12 to 24. This is satisfying, as $P=24$ indicates a daily periodicity for power consumption. Another observation is, in case of missing data, the AR model gives unreliable estimates for lower orders, which suggests, a correct choice of model order is crucial to compensate for the inaccuracies introduced by filling-in missing observations.

\subsection{Results on Traffic Data}

For the traffic dataset, the following configuration is used:
\begin{itemize}
	\item AR: $P=24$, $r_0 = 1$
	\item PMF: $d=20$, $\rho_u = 10^{-1}$, $\rho_v = 10^{-4}$
	\item CKF: $d=20$, $\nu_d = 10^{-6}$, $\nu_x = 10^{-6}$
	\item Naive MF: $d=20$, $\rho_u = 10^{-1}$, $\rho_v = 10^{-4}$
	\item FP-MF: $d=20$, $\rho_u = 10^{-1}$, $\rho_v = 10^{-4}$, $P=24$, $r_0 = 1$
	\item FT-MF: $d=20$, $\eps = 5 \times 10^{-2}$, $\rho_v = 10^{-4}$, $P=24$, $r_0 = 1$
	\item LN-MF: $d=20$, $\rho_v = 10^{-4}$, $P=24$, $r_0 = 1$
	\item ${\tt max\_ite} = 15$ for all algorithms.
\end{itemize}
In particular, $d$ has now increased from 5 to 20, accommodating for the dimensionality increase in the input data.

Once again, we consider structured and unstructured sparsity, as we did for the electricity data. The sparsity levels, arrival/departure rates, and the number of test sets are identical to what we used previously. In Figure \ref{fig:traf_mae_1} the results for unstructured sparsity is shown. In this case, once again the best results are given by the proposed methods, in particular, FT/ZT has the best predictions. On the other hand, Base and AR do not get much worse, compared to the analogous plot of Figure \ref{fig:elec_mae_1}. This shows that, performance of filling-based AR model does not only depend on the sparsity, but also the data itself. Interestingly, PMF and CKF are no longer competitive in this case. In Figure \ref{fig:traf_mae_2} we consider structured sparsity. This case set itself apart from all the others we considered so far. First, even for highly sparse inputs, the performance of the base estimator does not get worse at all. Figure \ref{fig:traf_mae_1} already shows that the sparsity does not have a very strong effect, even in adversarial case, so the results for structured noise are not surprising. Since this is true for the base predictor, the AR predictor remains competitive as well. In fact, here the fill step is good enough to alleviate missing data problem, so even if the data is sparse, the AR predictor can still provide accurate forecasts. If we instead fill missing entries with zeros, this apparent advantage of AR disappears. Nevertheless, when sparsity is less than a threshold the difference between AR and FP/FT/ZT is small. 

\begin{figure}[h]
\centering
\subfloat[]{\includegraphics[width = .5\columnwidth]{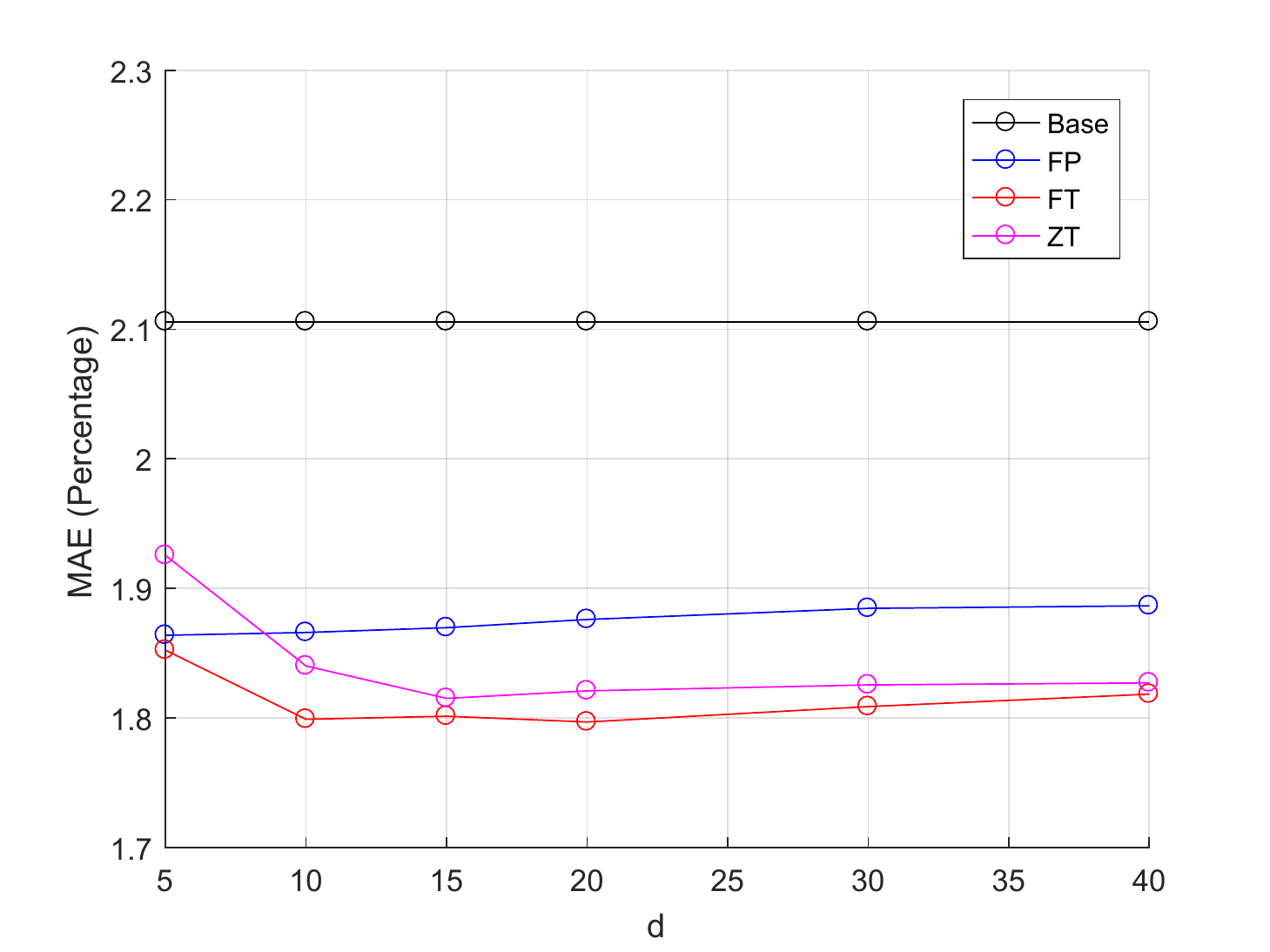}\label{fig:traf_ch_dim}}
\hfil
\subfloat[]{\includegraphics[width = .5\columnwidth]{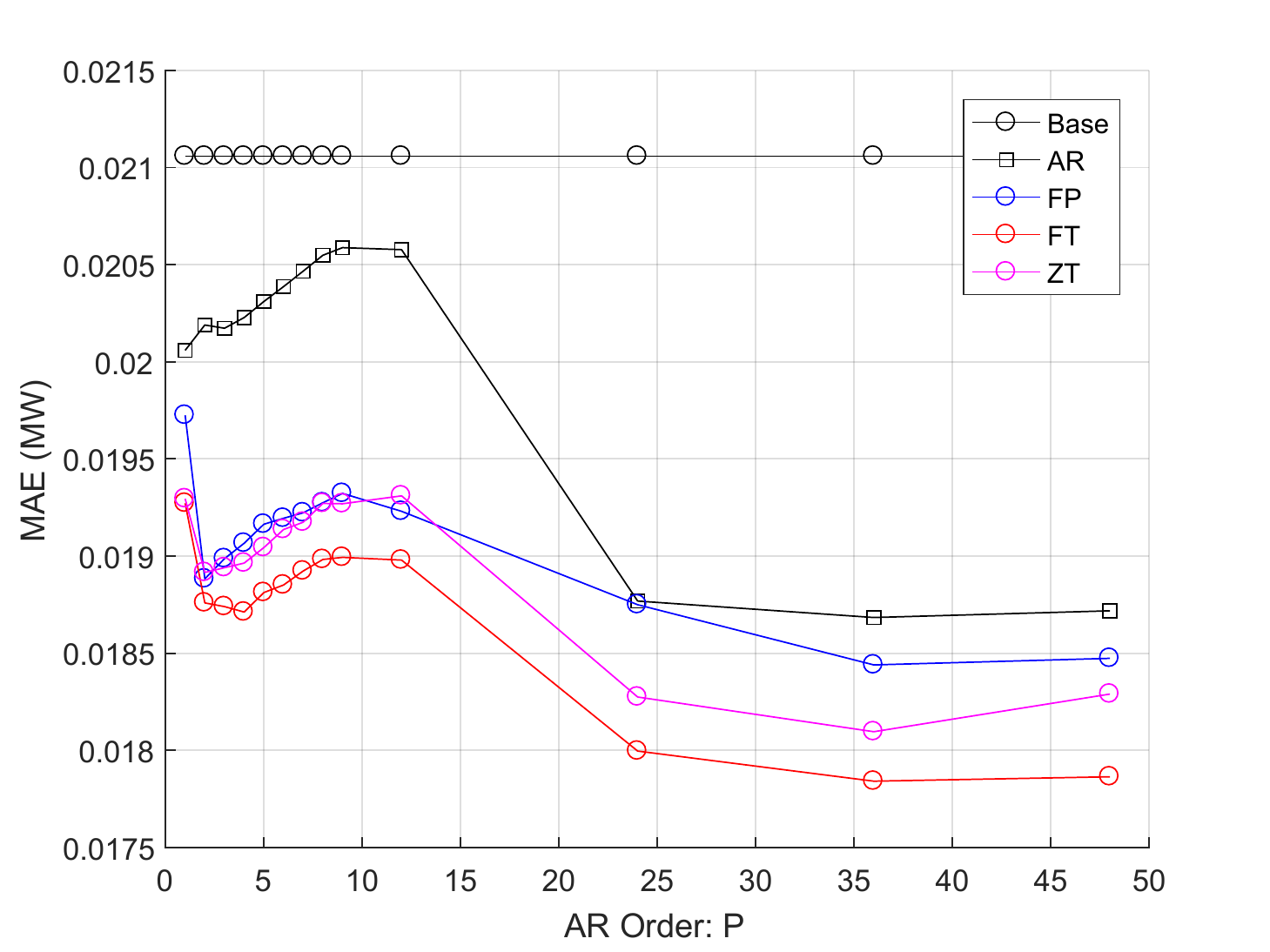}\label{fig:traf_ch_nord}}
\caption{Plot of prediction performances as a function of (a) rank and (b) AR order. Both plots obtained for traffic dataset with unstructured sparsity and NNZ = 50\%.}
\label{fig:traf_ch}
\end{figure}

Similar to electricity data, we now analyze the prediction performance as a function of rank and AR order in Figure \ref{fig:traf_ch}. In panel (a) we investigate the effect of latent dimensionality. Unlike the electricity data, choosing $d = 1,2$ results in unstable performance for FT and ZT; because for this dataset choosing such a low rank gives infeasible optimization problems. For this reason we sweep $d \in \{5, 10, 15, 20, 30, 40\}$ and observe once $d \geq 10$ all factorizations produce reliable results. Once again we note that ZT is more susceptible to error, compared to FT, when dimension is low, as the zero tolerance constraint is more restrictive. In panel (b) we show the effect of AR order. Here the results are similar to the electricity data, setting $P = 24$ yields good results for FP, FT, and ZT. Once again, a one day periodicity is expected, as the traffic intensity has a daily pattern, e.g. rush hours in the morning and evening.

\section{Conclusion}
\label{sec:conc}

We have considered the problem of forecasting future values of high dimensional time series. A high dimensional time series can be treated as a matrix, where each column denotes a cross-section, and when missing values are present, a low rank matrix factorization can be used as a building block for a forecaster. Based on this key idea, we proposed three methods which can perform matrix factorization in the online/streaming setting. These approaches differ by the type of regularization, and an important conclusion is, time varying regularization can be achieved through a constrained optimization problem, which is shown to have closed form solution. The matrix factorization component provides embeddings, which are then used to learn an AR model, which in turn forms the basis of forecasting. We have derived the optimum LMMSE estimator to find the AR coefficients, this estimator does not make any stationarity/Gaussianity assumptions, and only requires the first and second order statistics of the noise terms. Finally, we considered two real datasets on power demand and traffic, and showed that when missing values are present in the data, our methods will provide more reliable forecasts. As for future work, we plan to investigate the problem where entire columns of data are missing, which correspond system-wide blackouts, and also alternative factorization approaches that do not require any explicit regularization.

\appendices

\section{Fixed tolerance update: $\bU_t$} \label{app1}

At any given iteration-$i$ of the E-step in Algorithm \ref{alg:ft}, for a fixed $\bv_t^{(i)}$, we want to find
\begin{align}\label{eq:ft_U_opt}
	\bU_t^{(i)} = \arg\min_{\bU} \|\bU - \bU_{t-1}\|_F^2 ~~\text{s.t.}~~ \| \bx_t - \bU^\top \bv_t^{(i)} \|_2^2 \leq \eps ~.
\end{align}
Note the indexing, as $\bU_t^{(i)}$ is computed after $\bv_t^{(i)}$, as the latter appears before the former in the loop in Algorithm \ref{alg:ft}. To avoid clutter we will drop the time and iteration indices. We will also use $\bbU = \bU_{t-1}$ to distinguish the time indexing. From Eq. \eqref{eq:ft_U_opt} it is seen that the problem is strongly convex, and the feasible set is always nonempty. In fact, for any given $\bv_t^{(i)}$, the feasible set will have infinitely many elements. To see this, let $\eps=0$ and note that this gives $M_T$ linear equations in $d M_t$ unknowns. Since all of these solutions lie within the feasible set, the result follows. As a consequence, strong duality holds for this problem and we can characterize the solution via Karush Kuhn Tucker (KKT) conditions \cite{Boyd_2004}. For this problem these conditions read as
\begin{enumerate}
	\item $\nabla_{\bUst} \L = 0$
	\item $\| \bx - {\bUst}^\top \bv \|_2^2 \leq \eps$
	\item $\lamst \geq 0$
	\vspace{.05in}
	\item $\lamst \[ \| \bx - {\bUst}^\top \bv \|_2^2 - \eps \] = 0$
\end{enumerate}

Using the first condition, which needs to hold for primal feasibility, it is easily shown that the solution satisfies 
$$\bU = [\lambda^{-1} \bI + \bv \bv^\top]^{-1} (\lambda^{-1}\bbU + \bv \bx^\top) ~,$$
which can be re-written, using rank-1 version of matrix inversion lemma in Eq. \eqref{eq:MIL}, as
\begin{align} \label{eq:U_kkt_1}
	\bU = \bbU + \lambda \bv \bx^\top - \frac{\lambda}{1+\lambda \bv^\top \bv} (\bv \bv^\top) \bbU - \frac{\lambda^2}{1+\lambda \bv^\top \bv} (\bv \bv^\top) (\bv \bx^\top)
\end{align}
This alternative expression is useful, as the inverse term containing $\lambda$ disappears. Now, the third and fourth KKT conditions imply 
\begin{align} \label{eq:U_kkt_4}
\| \bx - \bU^\top \bv \|_2^2 = \eps	~.
\end{align}
This also satisfy condition number two. This means, the solution of the optimization problem has an approximation error that is equal to the maximum tolerance $\eps$. As $\bU$ is always updated after $\bv$ this means the training error of our algorithm at \emph{each} time step will be $\sqrt{\eps}$. This also suggests, we can find the value of Lagrange multiplier by plugging Eq. \eqref{eq:U_kkt_1} into Eq. \eqref{eq:U_kkt_4}. Now let the constants $c_1$-$c_4$ be as in Eq. \eqref{eq:ft_const}. We need to calculate the equation
$$\bv^\top \bU \bU^\top \bv - 2 \bx^\top \bU^\top \bv - (\eps - c_3) = 0 ~.$$
The first two terms evaluate as
\begin{align}
	&\bv^\top \bU \bU^\top \bv = c_4 +\lambda c_1 c_2 - \frac{\lambda}{1+\lambda c_2} c_2 c_4 - \frac{\lambda^2}{1+\lambda c_2} c_1 c_2^2 \nn \\
	 &+ \lambda c_1 c_2 + \lambda^2 c_2^2 c_3 - \frac{\lambda^2}{1+\lambda c_2} c_1 c_2^2 - \frac{\lambda^3}{1+\lambda c_2} c_2^3 c_3 \nn \\
	 &- \frac{\lambda}{1+\lambda c_2} c_2 c_4 - \frac{\lambda^2}{1 + \lambda c_2} c_1 c_2^2 + \frac{\lambda^2}{(1+\lambda c_2)^2} c_2^2 c_4 \nn \\
	 &+ \frac{\lambda^3}{(1+\lambda c_2)^2} c_1 c_2^3 - \frac{\lambda^2}{1+\lambda c_2} c_1 c_2^2 - \frac{\lambda^3}{1+\lambda c_2} c_2^3 c_3 \nn \\
	 &+ \frac{\lambda^3}{(1+\lambda c_2)^2} c_1 c_2^3 + \frac{\lambda^4}{(1+\lambda c_2)^2} c_2^4 c_3 \\
	& -2 \bx^\top \bU^\top \bv = -\frac{2 c_1}{1 + \lambda c_2} - \frac{2 \lambda c_2 c_3}{1 + \lambda c_2}
\end{align}
Multiplying with the denominator term $(1+\lambda c_2)^2$ and expanding the terms, the final expression is simply a second order polynomial
\begin{align}
	-\eps c_2^2 \lambda^2 - 2 \eps c_2 \lambda + (c_3 + c_4 - 2 c_1 - \eps) ~.
\end{align}
The roots are then given by the formula
\begin{align}
	-\frac{1}{c_2} \pm \frac{1}{\sqrt{\eps} c_2} \sqrt{c_3 + c_4 - 2c_1}
\end{align}
Since $c_2 = \|\bv\|_2^2 > 0$ the first term is negative. Then the third KKT condition implies, the second term above must be greater than the first term, and thus the polynomial always has one positive and one negative root. The update for the Lagrange multiplier in Eq. \eqref{eq:ft_lagm_U} now follows.

\section{Fixed tolerance update: $\bv_t$} \label{app2}

Similar to Appendix \ref{app1} we study iteration-$i$ of the E-step in Algorithm \ref{alg:ft}; but this time for a fixed $\bU_t^{(i)}$, we want to find
\begin{align}\label{eq:ft_v_opt}
	\bv_t^{(i+1)} = \arg\min_{\bv} \|\bv - \bv_{t-1}\|_2^2 ~~\text{s.t.}~~ \| \bx_t - \bU_t^{(i)} \bv_t \|_2^2 \leq \eps ~.
\end{align}
Unlike the case for $\bU_t$, it is not clear if the solution set is nonempty. In particular, note that in Appendix \ref{app1}, we showed that no matter how $\bv_t$ is chosen, the solution set always has infinitely many elements. For $\bv_t$ this is not true in general. To see this, consider the case where $\bU_t^\top \bv = \bx$ is an over constrained system of linear equations. The minimum error achievable in this case is given by the least squares solution as $\eps_\text{ls} = \bx^\top [\bI - \bU_t^\top(\bU_t^\top \bU_t)^{-1}\bU_t] \bx$. When $\eps_\text{ls} > \eps$ there are no solutions. However, if $\bU_t$ is updated before $\bv_t$, then it is guaranteed that the there is at least a single solution; because for the alternating optimization in Eq. \eqref{eq:ft_v_opt}, we are given that $\| \bU_t^{(i)} - \bv_t^{(i-1)} \|_2^2 < \eps$. As a result, $\bv_t^{(i-1)}$ is guaranteed to be in the feasible set. Therefore the solution set is nonempty and since the problem is strongly convex, strong duality holds. The KKT conditions mirror the previous one:
\begin{enumerate}
	\item $\nabla_{\bvst} \L = 0$
	\item $\| \bx - \bU^\top \bvst \|_2^2 \leq \eps$
	\item $\lamst \geq 0$
	\vspace{.05in}
	\item $\lamst \[ \| \bx - \bU^\top \bvst \|_2^2 - \eps \] = 0$
\end{enumerate}
Letting $\bbv = \bv_{t-1}$ the first condition yields
\begin{align}
	\bv = [\lambda^{-1} \bI + \bU \bU^\top]^{-1} [\lambda^{-1} \bbv + \bU \bx] ~.
\end{align}
This time, we cannot apply Sherman-Morrison identity, as $\bU$ is typically not rank-1. Instead, note that $\bU\bU^\top$ is in the positive semidefinite cone of symmetric matrices, and admits an eigendecomposition with nonnegative eigenvalues. Let's denote this by $\bU\bU^\top = \bQ \bPsi \bQ^\top$ , and define the constant vectors $\bc_1 = \bQ^\top \bU \bx$ and $\bc_2 = \bQ^\top \bbv$. Then it is straightforward to verify that
\begin{align}
	\bv = \bQ [\rho \bI + \bPsi]^{-1} \bc_1 + \bQ \rho [\rho \bI + \bPsi]^{-1} \bc_2
\end{align}
where we defined $\rho = 1/\lambda$; note the choice of notation here as $\rho$ is the regularizer, analogous to the one in FP matrix factorization.  Once again, from the fourth KKT condition we want to solve the equation
$$\bv^\top \bU \bU^\top \bv - 2 \bx^\top \bU^\top \bv - (\eps - c_3) = 0 ~,$$ 
where the first two terms are evaluated as
\begin{align}
	\bv^\top \bU \bU^\top \bv &= \sum_{i=1}^d \frac{\psi_i}{(\rho+\psi_i)^2} c_{1,i}^2 + \sum_{i=1}^d \frac{\rho^2 \psi_i}{(\rho+\psi_i)^2} c_{2,i}^2 \nn \\
	&\quad + 2 \frac{\rho \psi_i}{(\rho+\psi_i)^2} c_{1,i} c_{2,i} \nn \\
	- 2 \bx^\top \bU^\top \bv &= -2 \sum_{i=1}^{d} \frac{1}{\rho + \psi_i} c_{1,i}^2 -2 \sum_{i=1}^{d} \frac{\rho}{\rho+\psi_i} c_{1,i}c_{2,i}	
\end{align}
Substituting these into the equation and multiplying with the denominator term $\prod_{j=1}^{d} (\rho + \psi_i)^2$ we get the polynomial in Eq. \eqref{eq:ft_poly_v}.

\bibliographystyle{IEEEbib}
\bibliography{all_bib}

\end{document}